\documentclass[lettersize,journal]{IEEEtran}
\usepackage{amsmath,amsfonts}
\usepackage{algorithmic}
\usepackage{array}
\usepackage{textcomp}
\usepackage{stfloats}
\usepackage{url}
\usepackage{verbatim}
\usepackage{graphicx}
\def\BibTeX{{\rm B\kern-.05em{\sc i\kern-.025em b}\kern-.08em
    T\kern-.1667em\lower.7ex\hbox{E}\kern-.125emX}}
\usepackage{balance}
\usepackage{hyperref}
\usepackage{multirow}
\usepackage{orcidlink}
\usepackage{amsthm}
\usepackage{amssymb}
\usepackage{adjustbox}
\usepackage{booktabs}
\usepackage{amsmath}
\usepackage{subcaption}

\theoremstyle{definition}
\newtheorem{definition}{Definition}

\usepackage{xcolor}
\usepackage[textsize=tiny]{todonotes}

\begin{document}
\title{Protective Factor-Aware Dynamic Influence Learning for Suicide Risk Prediction on Social Media}
\author{IEEE Publication Technology Department
\thanks{Manuscript created October, 2020; This work was developed by the IEEE Publication Technology Department. This work is distributed under the \LaTeX \ Project Public License (LPPL) ( http://www.latex-project.org/ ) version 1.3. A copy of the LPPL, version 1.3, is included in the base \LaTeX \ documentation of all distributions of \LaTeX \ released 2003/12/01 or later. The opinions expressed here are entirely that of the author. No warranty is expressed or implied. User assumes all risk.}}

\markboth{Journal of \LaTeX\ Class Files,~Vol.~18, No.~9, September~2020}%
{How to Use the IEEEtran \LaTeX \ Templates}

\author{Jun Li~\orcidlink{0000-0002-2363-0641}, Xiangmeng Wang~\orcidlink{0000-0003-3643-3353}~\IEEEauthorrefmark{2}, Haoyang Li~\orcidlink{0000-0003-3152-5929}, Yifei Yan~\orcidlink{0000-0001-9014-6262}, Hong Va Leong~\orcidlink{0000-0001-7682-9032}, Ling Feng~\orcidlink{0000-0001-7378-4342}, Nancy Xiaonan Yu~\orcidlink{0000-0002-6371-2684}, Qing Li~\orcidlink{0000-0003-3370-471X}~\IEEEauthorrefmark{2}~\IEEEmembership{Fellow,~IEEE}

\IEEEcompsocitemizethanks{
\IEEEcompsocthanksitem 
Jun Li, X. Wang, H. Li, H. Leong, and Qing Li are with Department of Computing, The Hong Kong Polytechnic University, Hong Kong SAR, China (e-mail: hialex.li@connect.polyu.hk; 
\{xiangmengpoly.wang; haoyang-comp.li; hong.va.leong; qing-prof.li\}@polyu.edu.hk.
\IEEEcompsocthanksitem
Ling Feng is with the Department of Computer Science, Technology, Tsinghua University, Beijing 100190, China (e-mail: fengling@tsinghua.edu.cn).
\IEEEcompsocthanksitem
Y. Yan and N. Yu are with Department of Social and Behavioural Sciences, City University of Hong Kong, Hong Kong SAR, China (e-mail: yfyan8-c@my.cityu.edu.hk; nancy.yu@cityu.edu.hk).
}
\thanks{~\IEEEauthorrefmark{2}: Corresponding author}
}

\maketitle

\begin{abstract}
Suicide is a critical global health issue that requires urgent attention. Even though prior work has revealed valuable insights into detecting current suicide risk on social media, little attention has been paid to developing models that can predict subsequent suicide risk over time, limiting their ability to capture rapid fluctuations in individuals' mental state transitions. 
In addition, existing work ignores protective factors that play a crucial role in suicide risk prediction, focusing predominantly on risk factors alone.
Protective factors such as social support and coping strategies can mitigate suicide risk by moderating the impact of risk factors. Therefore, this study proposes a novel framework for predicting subsequent suicide risk by jointly learning the dynamic influence of both risk factors and protective factors on users' suicide risk transitions. 
We propose a novel Protective Factor-Aware Dataset, which is built from 12 years of Reddit posts along with comprehensive annotations of suicide risk and both risk and protective factors. We also introduce a Dynamic Factors Influence Learning approach that captures the varying impact of risk and protective factors on suicide risk transitions, recognizing that suicide risk fluctuates over time according to established psychological theories. Our thorough experiments demonstrate that the proposed model significantly outperforms state-of-the-art models and large language models across three datasets. In addition, the proposed Dynamic Factors Influence Learning provides interpretable weights, helping clinicians better understand suicidal patterns and enabling more targeted intervention strategies.
\end{abstract}

\section{Introduction}
Suicide issue remains one of society's most urgent challenges, with each case representing not only the loss of an individual life but also creating profound ripple effects across families, healthcare systems, and communities that persist for generations.
According to the World Health Organization, each year, approximately 727,000 individuals die from suicide around the world, particularly among young adults aged 15 to 29\footnote{https://www.who.int/news-room/fact-sheets/detail/suicide}. Detecting at-risk individuals is critical for preventing life-threatening outcomes by proactive psychological interventions.
Suicidal ideation detection (SID) is an emerging topic that determines whether the person has suicidal ideation.
Conventional approaches to SID 
often rely on professional resources such as questionnaires~\cite{fu2007predictive} and face-to-face consultation~\cite{scherer2013investigating}. However, these approaches are limited in their accessibility, particularly for individuals who are hesitant to reveal their thoughts or with limited access to mental health resources.

Fortunately, social media platforms provide relatively anonymous and open spaces for individuals to discuss mental health issues. 
Compared to conventional face-to-face approaches (e.g., consultations), social media platforms serve as a ``tree hole'', in which individuals can freely express their struggles without the fear of immediate judgment or stigma.
Unlike expensive psychological counseling methods, social media platforms are openly accessible, offering a low-cost and scalable alternative for monitoring SID.
More individuals are turning to social media platforms to share their personal struggles and emotional experiences. Therefore, recent research~\cite{karim2020social,ulvi2022social} has increasingly focused on SID methods that leverage social media data, bringing this topic to the forefront of social science and mental health research.

Existing research on SID methods can be roughly categorized as feature-based and deep learning algorithm-based methods.
Early approaches primarily focused on identifying relevant user features to assess risk factors, such as psychological lexicons (e.g., LIWC~\cite{pennebaker2001linguistic}) and fundamental linguistic attributes like n-grams and POS tags to predict the current suicide risk. 
For example, Sawhney et al. \cite{sawhney2018computational} extract
LIWC features, statistical features, and POS tag counts from tweets as the source of risk factors and employ logistic regression and ensemble classifiers to predict current suicidal ideation in tweets. 
With the advancement of deep learning, recent work has focused on using various deep learning methods to detect risk factors in posts.
For example, 
Daeun Lee et al.~\cite{DBLP:conf/kdd/LeeSJKH23} propose a multi-task learning framework that predicts the highest suicide risk level within the next several months based on the bipolar symptom (risk factors) of bipolar patients.
However, they ignore forecasting the suicide risk level of a subsequent post, which cannot enable more timely and targeted interventions before risk escalation occurs.
Despite the efforts, these existing works suffer from critical limitations: 
(1) 
They mainly focus on modeling risk factors in user posts, which are the factors that are explicitly related to negative expressions of users, e.g., expressions of hopelessness, isolation, or suicidal ideation. However, the protective factors, which are elements that protect individuals against suicide risk, have largely been overlooked.
These protective factors are crucial as they can counterbalance risk factors by acting to moderate the impact of risk factors on suicidality~\cite{johnson2011resilience}.
(2) 
They primarily assess current suicide risk by classifying risk levels in user content.
Current risk assessment, while useful, is inherently reactive and often limited in its ability to predict subsequent suicide risk transitions.
According to Fluid Vulnerability Theory \cite{rudd2006fluid}, suicide risk is not static but fluctuates rapidly over short periods.
This rapid fluctuation is difficult to capture through the methods in current suicide risk assessments and only provides infrequent glimpses into an individual's mental condition.
Instead, forecasting how risk levels will change in upcoming posts (i.e., subsequent risk prediction) would capture rapid factor fluctuations and achieve better SID performance. 

\begin{figure}[ht]
    \centering
    \includegraphics[width = 0.45\textwidth]{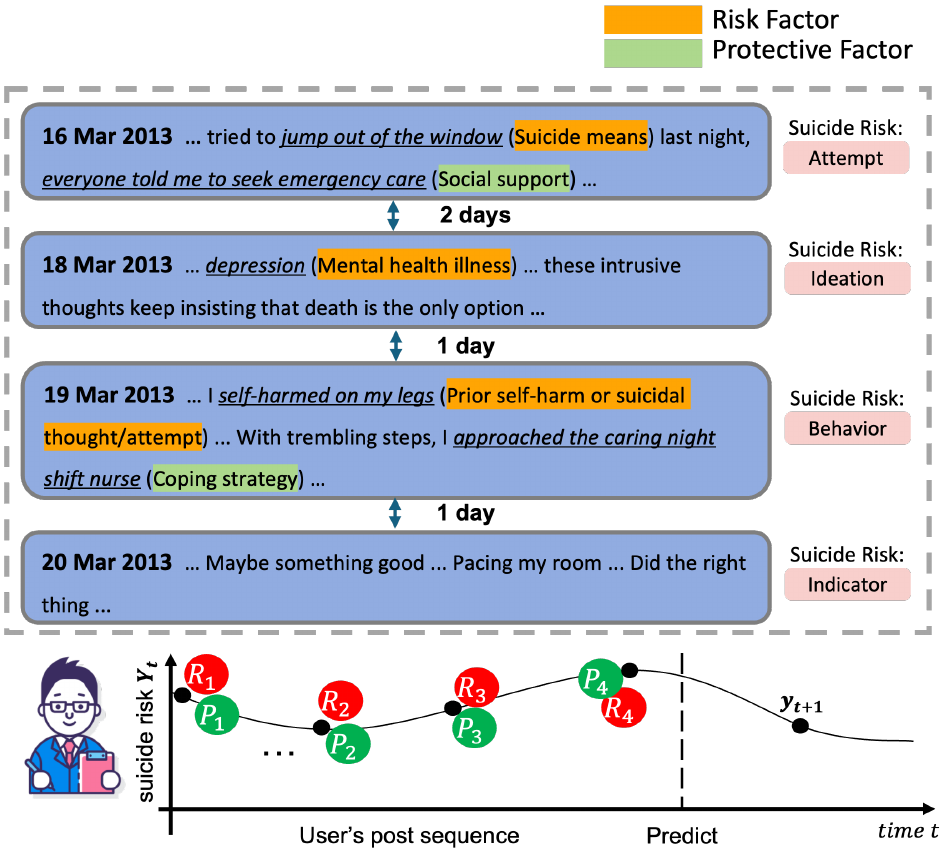}
    \caption{A toy example showing how both risk and protective factors are important for predicting a user's subsequent suicide risk, where red and green highlights indicate risk and protective factors.}
    \label{fig-example}
    \vspace{-0.15in}
\end{figure}

Protective factors, e.g., ``social support'', ``coping strategies'', and ``psychological capital'', form a distinct dimension from risk factors, which moderate the impact of risk factors on a suicide outcome~\cite{lundman2007psychometric}. They are equally important as risk factors in determining an individual's subsequent suicide risk level. 
In Figure~\ref{fig-example}, we use real-world posts acquired from Reddit to show the importance of protective factors. 
The user who reports a suicide means factor (with expressions: ``jump out of the window'') also receives ``social support'' (e.g., ``everyone told me to seek emergency care''). This protective factor facilitates the user's transition from the highest risk level (``Attempt'') to a lower risk level (``Ideation'') in the following post.
Similarly, the second-to-last post reveals ``coping strategy'' (e.g., ``approached the caring night shift nurse''), which appear to moderate the impact of risk factor ``prior self-harm or suicidal thought/attempt'' and facilitate the de-escalation from ``Behavior'' to ``Indicator'' suicide risk level in the final post. 
Due to the absence of protective factors, existing methods failed to capture the full dynamics of suicide risk transitions. Our work aims to fill this gap by incorporating protective factors, which enables us to learn how both risk and protective factors influence suicide risk changes.


Subsequent suicide risks reflect future suicidality of individuals, in contrast to current suicide risks.
Predicting those subsequent suicide risks is essential because individuals' suicide risks are inherently dynamic, fluctuating significantly over brief periods as indicated by Fluid Vulnerability Theory \cite{rudd2006fluid}.
Current suicide risks are static within a given user post context and are, therefore, unable to adapt to dynamic and time-sensitive suicide risk fluctuations.
Hence, exploring individuals' risk and protective factors that can lead to suicide ideation to predict future suicidality is crucial.
Therefore, this paper aims to predict the future suicidality of individuals based on the risk and protective factors revealed in their historical social media data, which has not been thoroughly investigated.




To bridge the gap, this work proposes to predict the subsequent suicide risk (i.e., future suicidality), given the risk and protective factors in users' historical social media data (cf. Figure~\ref{fig-example}).
We first create a novel SID dataset built from 12 years of Reddit posts and validated by well-trained researchers, with annotations for both risk factors and protective factors and subsequent suicide risk.
The dataset undergoes careful evaluation with high inter-annotator agreement, as detailed in Section~\ref{sec:evaluation of annotation}.
Unlike existing datasets that focus solely on risk factors, our dataset is the first to simultaneously include both protective factors and subsequent suicide risk annotations.
We conducted an empirical study on our proposed dataset (detailed in Section~\ref{sec:data analysis}), revealing that protective factors such as coping strategies are crucial determinants of subsequent suicide risk and frequently co-occur with risk factors.
Guided by our empirical observations, we propose a novel framework to learn risk and protective factors from post sequences and model their dynamic influence on suicide risk transitions for predicting subsequent suicide risk. 
Our approach differs from existing methods in two key aspects: First, we build Post Embedding and Temporal Context Modeling components to capture temporal patterns and contextual information from sequential posts, which are essential for modeling the evolving nature of mental health states. 
Second, we propose dynamic factor influence learning to capture how protective and risk factors exert varying degrees of influence on suicide risk transitions under different contextual conditions. 
Extensive experiments on three SID datasets demonstrate that our proposed model consistently outperforms large language models and state-of-the-art SID baselines, showing our model's superior performance and strong generalization capability.

\section{Related Work}

Social media platforms have emerged as tools for researchers to evaluate individuals' psychological well-being. The field has evolved significantly over time, with approaches becoming increasingly sophisticated in capturing the complex nature of suicidal expressions online. 
For instance, \cite{du2018extracting} utilizes convolutional neural networks (CNN) to classify the suicide/non-suicide tweets from social media by capturing the semantic information of each token in the tweets.
\cite{ji2018supervised} employs Long Short-Term Memory (LSTM) to capture sequential semantic information within the content to detect suicide ideation from both Twitter and Reddit datasets.
While these approaches effectively identified concerning content in isolated post, they lacked crucial user's context information derived from historical posts, limiting their ability to provide deeper insights into individuals' genuine suicidal expressions and reducing detection performance~\cite{venek2017adolescent}. For instance, phrases like ``I'm so depressed'' might represent genuine distress or casual exaggeration, while statements such as ``I want to die'' could indicate actual suicidal thoughts or merely express temporary frustration through figurative language. 
Recently, several studies propose to capture psychological linguistic features (i.e., emotions, symptoms, and risk factors) in users' posts, then model the temporal spectrum of these psychological features across time sequences to identify patterns indicative of suicide risk. 
For instance, drawing from psychological research on temporal emotional patterns, \cite{DBLP:conf/eacl/SawhneyJFS21} analyzes users' historical tweets to create time-aware emotional context of users, enhancing suicide risk detection accuracy on social media platforms.~\cite{lee2023towards} proposes a multi-task learning model that simultaneously predicts future suicidality and tracks bipolar disorder symptoms (i.e., manic mood, somatic complaints) in social media users.~\cite{cao2021learning} develops a two-stage approach that first uncovers users' inner real thoughts and emotion changes by analyzing correlations between public posts and hidden comments, then leverages these revealed inner states to detect suicide risk levels. \cite{sawhney2018computational} proposes a supervised approach that extracts multiple psychological linguistic features (i.e., LIWC features, POS counts, TF-IDF, and topic probabilities) from manually annotated tweets to train linear and ensemble classifiers for detecting suicidal ideation on Twitter. \cite{li2022suicide} proposes a dataset annotated by four well-trained experts with both risk factors and suicide risk labels, and develops a multi-task framework that jointly learns to identify risk factors and detect suicide risk within posts. However, a critical limitation still remains: \textit{they generally fail to incorporate protective factors, which play an important role in psychological assessment}. This oversight hinders accurate suicide risk evaluation and effective psychological intervention strategies, as understanding these protective factors is essential for comprehensive mental health support. Moreover, many people make impulsive suicide decisions with little planning, rather than carefully planned attempts. The time when someone is at the highest risk of taking action often lasts just minutes or hours~\cite{hawton2007restricting}. Given this brief critical period, existing methods either detect current suicide risk or predict future risk over periods of months, which, however, raises another limitation: \textit{they fail to predict the immediate subsequent risk that exists in the user's next post}. 
This gap in short-term predictive capability limits the effectiveness of timely interventions for high-risk individuals. Our work incorporates both protective factors and predicts users' immediate subsequent suicide risk, addressing the critical gap in short-term risk assessment for early intervention.

\section{Protective Factor-aware Data Collection}\label{sec:dataset}
We propose a Protective Factor-Aware Dataset (PFA) by collecting posts from 6,943 users published between June 15, 2010 and September 18, 2022 in the ``r/SuicideWatch'' subreddit. Following ethical guidelines, we use Reddit's public API to retrieve posts while removing all personally identifiable information to protect user privacy. We select users with sustained activity patterns (at least 7 posts within one week) to ensure rich temporal information, resulting in 237 users with 2,515 posts. Table~\ref{tab:descriptive_stats} shows the statistics of the dataset.

Three well-trained researchers in Psychology and Computer Science annotate the dataset with collaborative discussions to resolve disagreements. Unlike existing work, we incorporate three distinct label categories:
\textbf{A. Risk Factors}: We establish annotation criteria based on Li et al.~\cite{li2022suicide}, encompassing 15 key risk factors. Additionally, we identify four specific warning signs indicating imminent suicide risk (Prior self-harm/suicidal thought, Suicide means, Hopelessness, Emotion dysregulation) following Rudd et al.~\cite{rudd2006warning}.
\textbf{B. Protective Factors}: Following Wang et al. \cite{wang2022suicide}, we annotate five key protective factors (social support, coping strategy, psychological capital, sense of responsibility, and meaning in life) that represent core components of suicide resilience.
\textbf{C. Suicide Risk Levels}: We follow the Columbia Suicide Severity Rating Scale (C-SSRS)~\cite{posner2011columbia} to categorize suicide risks into indicator, ideation, behavior, and attempt levels.

\subsection{Evaluation of Annotation}\label{sec:evaluation of annotation}
\subsubsection{Cross validation} Our annotation process underwent rigorous consistency evaluation through the application of Fleiss' Kappa~\cite{fleiss1971measuring}, a statistical measure specifically designed to assess reliability across multiple annotators. The analysis revealed substantial inter-annotator agreement, with annotators achieving a high average Kappa coefficient of 0.84 for suicide risk level classification and 0.79 for both risk and protective factors annotation. These metrics indicate strong consensus among human evaluators and validate the reliability of our annotation framework.
\subsubsection{Comparison with Existing Datasets}
We compare our Protective Factor-Aware Dataset with four widely-used datasets from prior suicide research studies~\cite{gaur2019knowledge,li2022suicide,zirikly2019clpsych, ji2018supervised} as shown in Table~\ref{tab:dataset_comparison}. Our comparative analysis reveals several key distinctions that highlight the unique contributions of our dataset.
First, we observe that only two of the existing datasets incorporate risk factors, while the remaining datasets lack any suicidal factor annotations. More critically, none of the existing datasets simultaneously captures both the subsequent risk of suicide and protective factors, which is a significant limitation that restricts the comprehensive prediction of suicide risk. The inclusion of protective factors represents a particularly important advancement, as they are crucial for understanding resilience mechanisms. By incorporating both risk and protective dimensions, our dataset facilitates research into the complex interplay between risk and protective factors in suicidal behavior patterns, while also providing potential explanations for suicide trajectory evolution over time.

\begin{table}[!ht]
\caption{Statistics of dataset.}
\label{tab:descriptive_stats}
\centering
\begin{tabular}{lp{2cm}}
\hline
\textbf{Data Characteristics} & \textbf{Statistic} \\
\hline
Number of total posts & 2515 \\
Number of users & 237 \\
Distribution of suicide risk labels &  IN: 37.5\%; ID: 31.5\%; BR: 24.0\%; AT: 7.0\% \\
Number of risk factor categories & 19 \\
Number of protective factor categories & 5 \\
Posting interval (days) & 2.54 \\
\hline
\end{tabular}
\vspace{-0.1in}
\end{table}

\begin{table}[!ht]
\centering
\caption{Comparisons with existing datasets}
\label{tab:dataset_comparison}
\adjustbox{width=\columnwidth,center}{
\begin{tabular}{lccccc}
\toprule
 & \textbf{Ours} & \textbf{Gaur et al.} & \textbf{Li et al.} & \textbf{Zirikly et al.} & \textbf{Ji et al.} \\
 &  & \textbf{\cite{gaur2019knowledge}} & \textbf{\cite{li2022suicide}} & \textbf{\cite{zirikly2019clpsych}} & \textbf{\cite{ji2018supervised}} \\
\midrule
\textbf{Source} & Reddit & Reddit & Reddit & Reddit & Reddit and Twitter \\
\textbf{Time Duration} & 12 years & 11 years & 12 years & 10 years & - \\
\textbf{Subsequent Suicide Risk} & \checkmark & $\times$ & $\times$ & $\times$ & $\times$ \\
\textbf{Risk Factors} & \checkmark & $\times$ & \checkmark & $\times$ & $\times$ \\
\textbf{Protective Factors} & \checkmark & $\times$ & $\times$ & $\times$ & $\times$ \\
\textbf{Availability} & \checkmark & \checkmark & \checkmark & \checkmark & $\times$ \\
\bottomrule
\end{tabular}
}
\vspace{-0.25in}
\end{table}

\begin{table*}[ht]
\caption{Specifications of factor categories and suicide risk levels.}
\label{tab:specifications of annotation}
\centering
\small
\begin{tabular}{p{2cm}p{5cm}p{10cm}}
\hline
\textbf{Main Category} & \textbf{Sub-Category} & \textbf{Definition} \\
\hline
\multirow{19}{*}{Risk factors} & Mental health illness (MHI) & Existing mental disorders (e.g., depression, personality disorders). \\
 & Physical health/characteristic (PH) & Physical health issues (e.g., COVID-19, obesity/underweight). \\
 & Substance use (SU) & The uncontrollable use of drugs/alcohol/tobacco. \\
 & Hopelessness (HL) & Feelings of hopelessness, or feeling trapped/stuck. \\
 & Emotion dysregulation (ED) & Emotion regulation difficulties (e.g., uncontrolled anxiety, anger). \\
 & Low self-esteem (LS) & Self-negative feelings (e.g., worthlessness, being a burden, self-hate). \\
 & Poor school performance (PSP) & Low school performance (e.g., failing tests, bad grades). \\
 & Low socio-economic status (LSS) & Unemployment status, poverty, and homeless, etc. \\
 & Interpersonal violence (IV) & Violence or assault, occurring outside home settings. \\
 & Prior self-harm or suicidal thought/attempt (PSST) & Past, resolved, previous self-harm and suicidal thought/plan/attempt. \\
 & Poor social support (PSS) & Lack friends, loneliness, being isolated/rejected/neglected/abandoned, etc. \\
 & Interpersonal difficulty (ID) & Having difficulty making new friends, socialization, etc. \\
 & Dysfunctional family (DF) & Family-related issues that leads to negative impacts on the individual. \\
 & Exposure to others' suicide (EOS) & Mentioning or describing others' suicidal thoughts, attempt or death. \\
 & Stressful life event (SLE) & Events that pose challenges but do not lead to traumatic responses. \\
 & Traumatic experience (TE) & Event(s) that is overwhelming an individual’s ability to cope. \\
 & Cognitive deficit (CD) & Having difficulty in cognitive abilities. \\
 & Suicide means (SM) & Description of potential means or access to that means. \\
 & Sexual orientation related issues (SORI) & Having gender/sexual disorder, same-sex relationship, etc. \\
\hline
\multirow{5}{*}{Protective factors} & Social support (SS) & Support from family, partners, friends, healthcare professionals. \\
 & Coping strategy (CS) & Activities individuals engage in to deal with stressful situations.\\
 & Psychological capital (PC) & Individual's positive psychological state of development. \\
 & Sense of responsibility (SR) & Awareness of responsibility to one's own health/survival and to others. \\
 & Meaning in life (ML) & Involves cognitive component, motivational component and affective component. \\
\hline
\multirow{4}{*}{Suicide risk} & Indicator (IN) & Post contains no explicit mentions of suicide. \\
& Ideation (ID) & Post contains explicit suicidal expressions but no suicidal plan. \\
& Behavior (BR) & Post contains explicit suicidal expressions and self-harm/suicidal plan. \\
& Attempt (AT) & Post contains explicit mentions of recent/past suicidal attempts. \\
\hline
\end{tabular}
\vspace{-0.2in}
\end{table*}

\section{Data Analysis}\label{sec:data analysis}
In this section, we analyze our protective factor-aware dataset introduced in Section~\ref{sec:dataset} to answer the following questions:
\begin{itemize}
    \item \textit{Q1. To what extent do risk and protective factors impact subsequent suicide risk?}
    \item  \textit{Q2. To what extent are risk and protective factors correlated?}
\end{itemize}


To answer Q1, we classify our dataset into low-risk (IN, ID) and high-risk (BR, AT) suicide groups based on subsequent suicide risk. We then apply Chi-Square tests to identify factors showing statistically significant differences between groups. We show the Chi-Square test results in Table~\ref{tab:factors_test_statistics}. Five factors significantly distinguish between high and low suicide risk groups: ``suicide means'' (SM), ``prior self-harm or suicidal thought/attempt'' (PSST), ``coping strategy'' (CS), ``traumatic experience'' (TE), and ``physical health/characteristic'' (PH). These findings align with clinical studies identifying SM, PSST, TE, and PH as major risk factors, while CS serves as a strong protective factor for relieving suicidal risk~\cite{amare2018prevalence,amiri2022prevalence}. Surprisingly, ``social support'' (SS) shows lower statistical significance compared to dominant factors, contrasting with prior studies that highlight its protective role~\cite{harris2019factors, henderson2015responses}. This may reflect the unique social media context where users already receive baseline social support through online interactions. Additionally, ``mental health issues'' (MHI) demonstrate lower significance than expected, suggesting that clinical MHI alone may not be the sole indicator of suicide risk and that multi-level indicators including protective factors should be considered for more precise detection.
\begin{table}[!htbp]
\centering
\caption{$\chi^2$ values for factors differentiating between high and low suicide risk groups. ** indicates the $p$-value of the factors is less than 0.05 ($p<0.05$), which is considered highly statistically significant.}
\begin{tabular}{lclclc}
\hline
\textbf{Factors} & $\chi^2$ & \textbf{Factors} & $\chi^2$ & \textbf{Factors} & $\chi^2$ \\
\hline
SM & 64.96** & ED & 1.02 & SS & 0.13 \\
PSST & 33.28** & IV & 0.95 & CD & 0.08 \\
CS & 6.06** & ID & 0.66 & LS & 0.06 \\
TE & 4.36** & ML & 0.64 & LSS & 0.04 \\
PH & 3.92** & SORI & 0.62 & DF & 0.02 \\
HL & 2.26 & EOS & 0.50 & MHI & 0.01 \\
PC & 1.94 & PSP & 0.42 & SU & 0.002 \\
PSS & 1.55 & SLE & 0.23 & & \\
SR & 1.37 & & & & \\
\hline
\end{tabular}
\label{tab:factors_test_statistics}
\vspace{-0.2in}
\end{table}

\begin{figure}[ht]
    \centering
    \includegraphics[width = 0.48\textwidth]{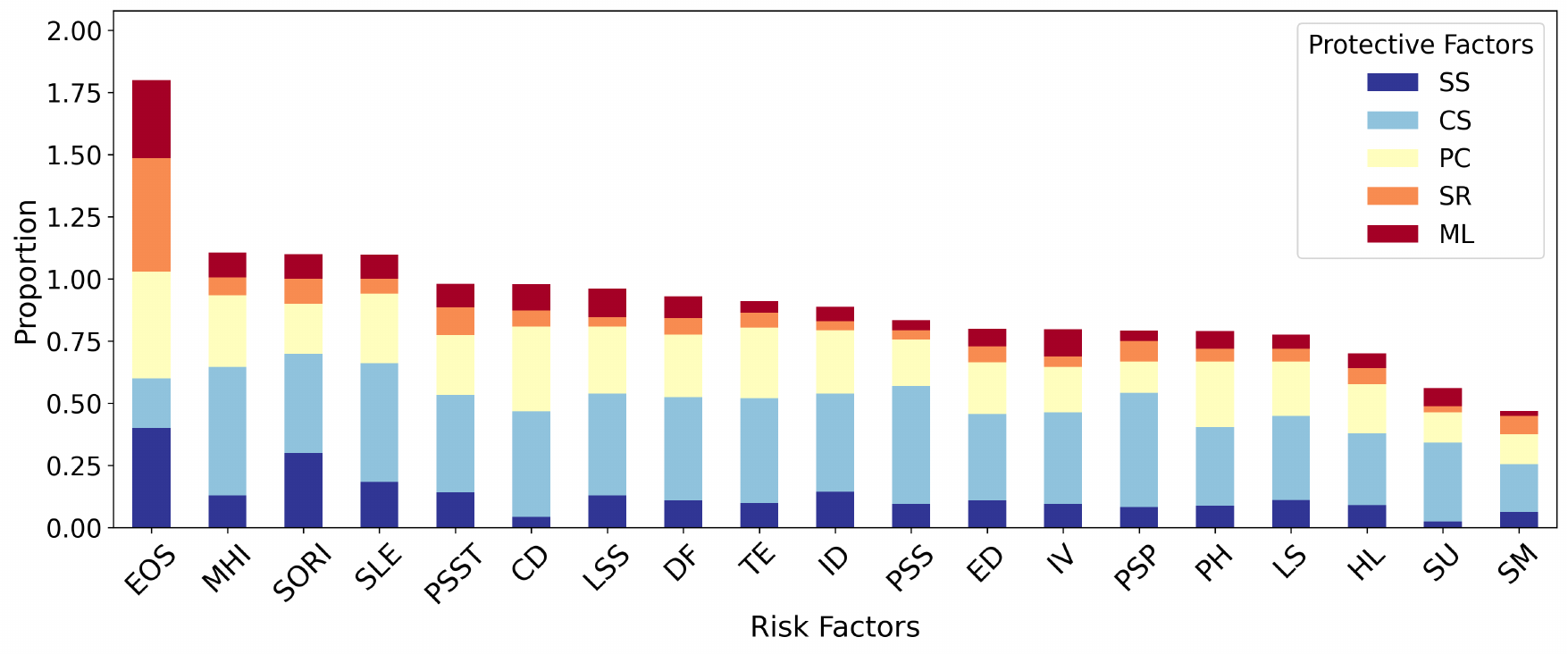}
    \caption{Distribution of co-occurrence $P_{ij}$ across different risk-protective factor pair $i-j$.}
    \label{fig-protective factors distribution}
    \vspace{-0.2in}
\end{figure}

Individuals at risk of suicide typically exhibit various risk factors while simultaneously possessing protective factors, as indicated by the existing literature~\cite{o2014psychology}.
Therefore, we are interested in answering Q2 - how risk and protective factors are correlated.
We define a co-occurrence $P_{ij}$ as the likelihood that a particular risk factor $i$ co-occurs with a protective factor $j$. Formally,
\begin{equation}
\begin{aligned}
\label{eq:protective_proportion}
P_{ij} = \frac{Count(RF_i \cap PF_j)}{Count(RF_i)}
\end{aligned}
\end{equation}
where $RF_i$ denotes the $i$-th risk factor and $PF_j$ represents the $j$-th protective factor. $Count(RF_i)$ represents the total number of users with risk factor $RF_i$, while $Count(RF_i \cap PF_j)$ counts users who simultaneously exhibit both risk factor $RF_i$ and protective factor $PF_j$. This metric enables us to examine the co-occurrence patterns of protective factors across different risk factor categories. 
We show the distribution of $P_{ij}$ on each risk-protective factor pair in Figure~\ref{fig-protective factors distribution}.
As illustrated in Figure~\ref{fig-protective factors distribution}, our analysis reveals distinct patterns in associations between risk factors and protective factors. Specifically, risk factors such as ``exposure to others' suicide'' (EOS) and ``mental health issues'' (MHI) demonstrate high rates of co-occurrence with protective factors, potentially because these individuals suffering from EOS and MHI may selectively receive more intensive support systems while simultaneously developing adaptive coping capacities in response to their high-risk exposure.
In contrast, ``suicide means'' (SM) and ``substance use'' (SU) are less correlated with protective factors, suggesting these individuals with SM and SU tend to receive less protective support.
This phenomenon may be attributed to the fact that people who mention suicide means or substance use may be especially vulnerable and less likely to seek help. In such cases, protective factors are less effective in helping moderate suicidal thoughts. Therefore, it is important to strengthen the influence of protective factors, especially for these high-risk groups.


\section{Task Formulation}\label{sec:task formulation}


Risk and protective factors fluctuate in reaction to environmental contingencies and/or internal experiences over time; thus, suicide risk constantly changes~\cite{bryan2020nonlinear}. 
As users' posts from different post sequences display diverse risk and protective factors and suicide risk levels, it is critical to predict subsequent suicide risk based on users' previous post sequences.
Our task is to capture those dynamic suicidality changes within multiple user post sequences to better predict users' subsequent suicide risks.
Following~\cite{DBLP:conf/kdd/LeeSJKH23}, we first create multiple post sequences for each user. Specifically, we set a window by selecting the past $l$ number of posts as given post sequences to predict the suicide risk of subsequent posts. 
In particular, 
we implement a sliding window over chronological posts for each user, shifting the window by a single post each time to generate multiple consecutive post sequences. In Section~\ref{sec:section7}, we 
analyze the impact of $l$ (i.e., the number of historical posts) and determine the optimal $l$ needed to achieve accurate suicide risk prediction while maintaining computational efficiency. 

Having established those temporal-aware users' post sequences, this work aims to capture the embedded dynamic suicidality changes under post sequences for subsequent suicide risk prediction. Formally, 



\begin{definition}[Subsequent Suicide Risk Prediction]
\label{task formulation}
     Given a user's dynamic post sequence, $P_u=\{p_{1},p_{2},...,p_{t}\}$ over time steps $T = \{1,...,t\}$, where the user $u \in U=\{u_1, u_2,...,u_n\}$ and $U$ is the total set of users. 
     We denote the ground truth of suicide risk, risk factors, and protective factors as $Y^{sr}$, $Y^{rf}$, and $Y^{pf}$. 
     Our task is to (1) classify multiple risk factors $Y^{rf} = \{y^{rf}_1, y^{rf}_2, ..., y^{rf}_M\}$ 
     and protective factors $Y^{pf} = \{y^{pf}_1, y^{pf}_2, ..., y^{pf}_K\}$ that exist in each post $p_{t}$, where $M$ and $K$ are the numbers of risk and protective factors respectively. (2) predict users' subsequent suicide risk $y^{sr}_{t+1} \in Y^{sr}=\{\text{``IN''},\text{``ID''},\text{``BR''},\text{``AT''}\}$ at subsequent time step $t+1$ given risk factors $Y^{rf}$ and protective factors $Y^{pf}$. 
\end{definition}



\section{Methodology}

\begin{figure*}[ht]
    \centering
    \includegraphics[width =0.85\textwidth]{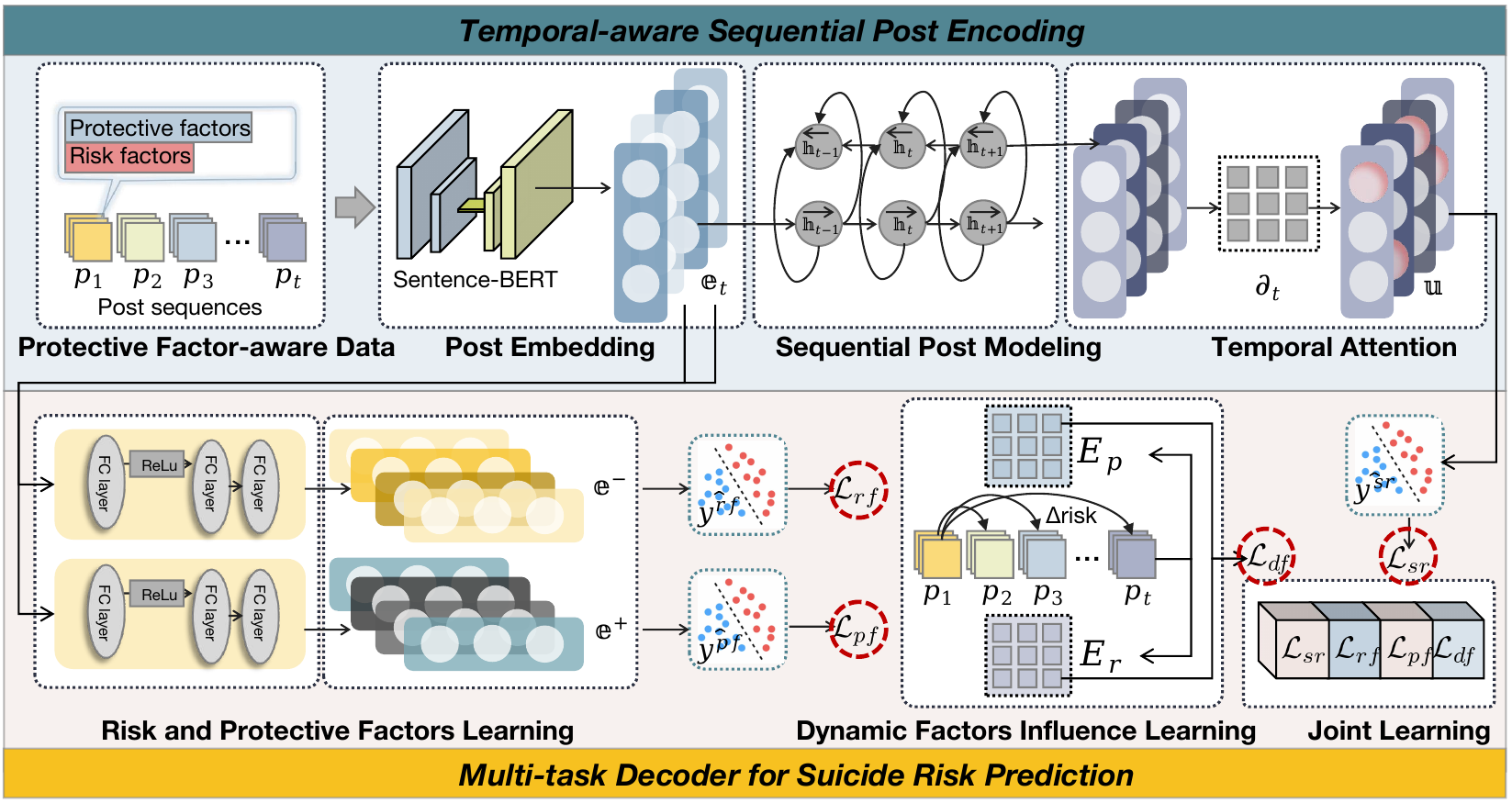}
    \caption{The overall architecture of the proposed model for subsequent suicide risk prediction. Our framework consists of two main components: (1) Temporal-aware Sequential Post Encoding, which employs Sentence-BERT for post embedding, followed by Sequential Post Modeling (BiLSTM) and Temporal Attention to capture temporal patterns and contextual information; (2) Multi-task Decoder for Suicide Risk Prediction, which simultaneously learns risk and protective factors from individual posts through dedicated MLPs, models their dynamic influence on suicide risk transitions through Dynamic Factors Influence Learning (learning the influence of both types of factors according to their effectiveness: $E_p$ and $E_r$), and integrates all components via joint learning to predict subsequent suicide risk levels.}
    \label{fig-framework}
    \vspace{-0.2in}
\end{figure*}

We now formally introduce our proposed method to accomplish the task as in Definition.~\ref{task formulation}, with the overall framework shown in
Figure~\ref{fig-framework}. 

\subsection{Post Embedding}
Each user's post contains valuable semantic information that could indicate mental health states and suicide risk~\cite{de2013predicting}. Given a user's dynamic post sequence, $P_u=\{p_{1},p_{2},...,p_{t}\}$, we embed each post by using our \textit{Post Embedding} module to capture the semantic information in it. 
Specifically, we employ Sentence-BERT (SBERT)~\cite{reimers2019sentence}, which has demonstrated effectiveness in representing user's posts~\cite{DBLP:conf/kdd/LeeSJKH23} and detecting mood state changes~\cite{azim2022detecting}. SBERT is an extension of the pre-trained BERT model that preserves semantic relationships through siamese and triplet network structures, making it well-suited for capturing the nuanced mood and psychological state expressions in social media posts.  
Formally,
the embedding of each post $p_t$ is obtained by using the Sentence-BERT:
\begin{equation}
\begin{aligned}
\label{eq:post embedding}
\mathbf{e}_t=SBERT(p_t)
\end{aligned}
\end{equation}
where $\mathbf{e}_t$ is the post embedding and $SBERT(\cdot)$ represents the operator of embedding by Sentence-BERT model.


\subsection{Temporal Context Modeling}

To capture the dynamic nature of risk factors, protective factors, and suicide risk levels that fluctuate over time~\cite{oliffe2012you}, we propose a \textit{Temporal Context Modeling} to model their temporal evolution within user posts. First, we utilize \textit{Sequential Post Modeling} to capture the sequential information of each post while considering the user's posting context. Then we employ the \textit{Temporal Attention} module to pay more attention to the essential factors that influence the model decision on subsequent suicide risk. 
These modules enable our model to track the trajectory of a user's mental state, capturing both sequential information and factor importance in post sequences.


\subsubsection{Sequential Post Modeling} 
To encode the context of each post sequence, we implement a BiLSTM~\cite{hochreiter1997long} network that captures sequential information effectively by building long-term dependencies from both directions.
Specifically, we feed each post embedding $\mathbf{e_t}$ acquired from Eq.~\eqref{eq:post embedding} into a Bidirectional LSTM, which processes the sequence in both forward and backward directions. The sequential representation $\mathbf{h_t}$ is formed by concatenating the hidden states from both passes, capturing sequential information from the entire post sequence.
Formally, given a user $u$'s semantic post embedding
$\mathbf{e}_t$ at time $t$ from the \textit{Post Embedding} module, 
the sequential embedding $\mathbf{h}_t$ of $u$ given $\mathbf{e}_t$
is acquired by the following equation: 
\begin{equation}
\begin{aligned}
\label{eq:context embedding}
\overrightarrow{\mathbf{h}_t} &= \text{LSTM}(\mathbf{e}_t, \overrightarrow{\mathbf{h}}_{t-1}) \\
\overleftarrow{\mathbf{h}_t} &= \text{LSTM}(\mathbf{e}_t, \overleftarrow{\mathbf{h}}_{t+1}) \\
\mathbf{h}_t &= [\overrightarrow{\mathbf{h}_t}, \overleftarrow{\mathbf{h}_t}]
\end{aligned}
\end{equation}
where $\overrightarrow{\mathbf{h}_t}$ and $\overleftarrow{\mathbf{h}_t}$ denote the post sequence at time $t$ processed from left to right and from right to left separately in BiLSTM.  \text{LSTM} is the operator of the LSTM neural network~\cite{hochreiter1997long}. 
$\mathbf{h}_t\in \mathbb{R}^d$ is sequential representation of each post embedding $\mathbf{e}_t$; $d$ is the dimension of $\mathbf{h}_t$.
After the \textit{Sequential Post Modeling}, we obtain a sequential representation of each post that incorporates its long-term history.

\subsubsection{Temporal Attention} 
While BiLSTM in \textit{Sequential Post Modeling} module captures sequential dependencies, not all posts contribute equally to the user's current mental state.
This is because risk and protective factors in post sequences exhibit temporal dynamics. Traditional attention approaches fail to capture this dynamic nature. They overlook how the timing of each factor affects subsequent suicide risk differently.
The time intervals between factors appearing in posts vary considerably. To interpret the temporal effect of these factors, we consider how the duration between factor occurrences influences their impact on subsequent suicide risk. Therefore, given the time interval between posts within the post sequence, we employ temporal attention~\cite{lee2023towards} to dynamically weigh the importance of different content across the dynamic post sequence. 
We first calculate the attention score $\mathbf{a}_t$ by using the attention mechanism:

\begin{equation}
\mathbf{a}_t = \frac{\exp(f_1(\boldsymbol{\delta}_t))}{\sum_{i=1}^{T} \exp(f_1(\boldsymbol{\delta}_t))}, \text{ where }
\boldsymbol{\delta}_t = \sigma(\theta - \mu \Delta t)\mathbf{h}_t
\label{eq: attention score}
\end{equation}
where $f_1(\cdot)$ is full-connected layer with tanh activation. $\mathbf{h}_t$ is the sequential representation that we learned from Eq.~\eqref{eq:context embedding}. $\boldsymbol{\delta}_t$ represents the hidden state incorporating temporal decay. $\Delta t$ is time interval between posts. $\theta, \mu$ are learnable parameters. $\sigma$ is sigmoid function.

We then apply the attention score $\mathbf{a}_t$ acquired from Eq.~\eqref{eq: attention score} to the sequential representation $\mathbf{h}_t$ as below:
\begin{equation}
\mathbf{u} = \sum_{t=1}^{T} \mathbf{a}_t \mathbf{h}_t
\label{eq:attention}
\end{equation}
where $\mathbf{u}$ is temporal attention-aware representation of post sequence.
Till now, this module generates $\mathbf{u}$ by paying attention to the posts that are most indicative of suicide risk evolution over time.

\subsection{Multi-task Decoder for Suicide Risk Prediction}

We propose a \textit{Multi-task Decoder for Suicide Risk Prediction} to capture the dynamic evolution of user mental patterns through temporal modeling. The decoder consists of four key components.



\subsubsection{Risk and Protective Factors Learning}

According to the buffering hypothesis~\cite{johnson2011resilience}, protective factors should be treated as existing on a separate dimension from risk factors. This means protective factors operate independently rather than being the reverse of risk factors. Therefore, we implement a factor encoder to separately learn the risk and protective factors from each post. This approach allows us to capture the unique characteristics and contributions of both factor types in our predictive model. Specifically, using the post embedding $\mathbf{e_t}$ obtained from the \textit{Post Embedding} module in Eq.~\eqref{eq:post embedding} for each post $p_t$, our model calculates the logits for both risk factors and protective factors classification as follows:

\begin{equation}
\hat{y}^{rf} = f_4(\mathbf{e}^-) ,\text{ where }\mathbf{e}^- = ReLU(f_2(ReLU(f_3(\mathbf{e_t}))))
\label{eq:risk factors predictions}
\end{equation}



\begin{equation}
\hat{y}^{pf} = f_7(\mathbf{e}^+) ,\text{ where }\mathbf{e}^+ = ReLU(f_5(ReLU(f_6(\mathbf{e_t}))))
\label{eq:protective factors predictions}
\end{equation}
where $\mathbf{e}^-$ and $\mathbf{e}^+$ are risk factor embedding and protective factor embedding. $f_2(.)$, $f_3(.)$, $f_4(.)$, $f_5(.)$, $f_6(.)$ and $f_7(.)$ are full-connected layers. $ReLU$ is an activation function. $\hat{y}^{rf}$ and $\hat{y}^{pf}$ are predicted logits of risk factors and protective factors. 
For the factor identification task, we can treat it as a multi-label classification. Therefore, the objective functions of learning risk factors and protective factors are given as follows:

For risk factors:
\begin{equation}
\mathcal{L}_{rf} = -\sum_{j=1}^{M}[y_{j}^{rf} \log(\hat{y}_{j}^{rf})) + (1-y_{j}^{rf})\log(1-\hat{y}_{j}^{rf})]
\label{eq: loss function for learning risk factors}
\end{equation}


For protective factors:
\begin{equation}
\mathcal{L}_{pf} = -\sum_{j=1}^{K} [y_{j}^{pf} \log(\hat{y}_{j}^{pf}) + (1-y_{j}^{pf})\log(1-\hat{y}_{j}^{pf})]
\label{eq: loss function for learning protective factors}
\end{equation}
where $M$ and $K$ are the number of risk and protective factor labels, respectively. $y_{j}^{rf}$ and $y_{j}^{pf}$ are the ground truth labels\footnote{The ground truth labeling process is described in Section~\ref{sec:dataset}}.
$\hat{y}_{j}^{rf}$ and $\hat{y}_{j}^{pf}$ are the predicted logits derived from Eq.~\eqref{eq:risk factors predictions} and Eq.~\eqref{eq:protective factors predictions}, respectively.

\subsubsection{Dynamic Factors Influence Learning}
In subsequent suicide risk prediction, the impact of protective and risk factors varies significantly depending on the context of the post sequence of different users. 
Notably, a single crucial protective factor can dominate and significantly reduce subsequent suicide risk, even in the presence of multiple risk factors. This dynamic interplay challenges traditional models that treat all factors with equal importance.
To address this challenge, we propose a dynamic factor influence approach that can identify and learn from dominant factors. Our approach consists of three key components: (1) \textit{Effectiveness measurement} aims to measure the effectiveness of factors in changing the suicide risk trajectory. (2) \textit{Factor-state alignment learning} aims to learn associations between factors and the user’s post sequence. (3) \textit{Dynamic factor integration} aims to drive the model to prioritize learning the effective factors identified through \textit{Effectiveness measurement}.

\noindent \textbf{Effectiveness Measurement}. First, we define how to measure the effectiveness of different factors in influencing suicide risk transitions. Let $\Delta risk = y_{t+1} - y_t$ represent the change in suicide risk between consecutive time steps. We introduce Bernoulli-based functions to capture factor effectiveness. 

For protective factors, as protective factors could lower the risk of users' suicide attempts in the future, we thus define the Bernoulli function of protective factors as:
\begin{equation}
E_{p} = \begin{cases} 
1 & \text{if } \Delta \text{risk} < 0 \\
0 & \text{otherwise} 
\end{cases}
\label{eq:E_p}
\end{equation}

Similarly, we define the Bernoulli function to denote the probability that risk factors elevate suicide risk:

\begin{equation}
E_{r} = \begin{cases} 
1 & \text{if } \Delta \text{risk} > 0 \\
0 & \text{otherwise} 
\end{cases}
\label{eq:E_r}
\end{equation}
Thus, we can drive the model to learn how each type of factor influences suicide risk transitions based on the defined effectiveness measures.

\noindent \textbf{Factor-state Alignment Learning}. To identify which factors are most relevant to a user, we develop an alignment learning method that measures the association between dynamic post sequence and potential influencing factors. As shown in Figure~\ref{fig-framework}, for each dynamic post sequence embedding $\mathbf{u}$ acquired from Eq.~\eqref{eq:attention}, we define the association strength between post sequence and two types of factors through the following functions:
\begin{equation}
S_{p} = \frac{\sum_{t=1}^Texp(sim(\mathbf{u}, \mathbf{e}_{t}^+)/\tau)}{\sum_{t=1}^Texp(sim(\mathbf{u}, \mathbf{e}_{t}^+)/\tau) + \sum_{t=1}^Texp(sim(\mathbf{u}, \mathbf{e}_{t}^-)/\tau)}
\label{eq:S_p}
\end{equation}

\begin{equation}
S_{r} = \frac{\sum_{t=1}^Texp(sim(\mathbf{u}, \mathbf{e}_{t}^-)/\tau)}{\sum_{t=1}^Texp(sim(\mathbf{u}, \mathbf{e}_{t}^+)/\tau) + \sum_{t=1}^Texp(sim(\mathbf{u}, \mathbf{e}_{t}^-)/\tau)}
\label{eq:S_r}
\end{equation}
where $T=\{1,\cdots,t\}$ denotes time steps. The temporal attention-aware embedding of post sequence $\mathbf{u}$ is acquired from Eq.~\eqref{eq:attention}. Protective factor embeddings $\mathbf{e}^{-}$ and risk factor embeddings $\mathbf{e}^{+}$ are obtained from Eq.~\eqref{eq:risk factors predictions} and Eq.~\eqref{eq:protective factors predictions}, respectively
Here, $sim(\cdot,\cdot)$ represents the cosine similarity. $S_p$ and $S_r$ quantify the alignment strength, and $\tau$ controls the sensitivity of the alignment measurement. This alignment learning ensures suicide risk predictions are dynamically informed by the most relevant influencing factors. By explicitly modeling the alignment between post sequence and contributing factors, our model achieves suicide risk prediction that holistically considers both posting content and underlying suicidal factors.

\noindent \textbf{Dynamic Factor Integration}. 
To capture the dynamic nature of factor influence, we propose a dynamic weighting mechanism that learns dynamic weights of factors through the measurements of their effectiveness: 
\begin{equation}
\begin{aligned}
\mathcal{L}_{df} = \frac{1}{\left|V\right|}\sum_{j=1}^{\left|V\right|} \Big[\frac{1}{2}\big(&-E_{p}log(S_{p}) - (1-E_{p})log(1-S_{p}) \\ 
&-E_{r}log(S_{r}) - (1-E_{r})log(1-S_{r})\big)\Big]
\end{aligned}
\label{eq: loss function for dynamic factor integration}
\end{equation}
where $V = \left\{ i \in I \colon E_p^i = 1 \vee E_r^i = 1 \right\}$.
$E_p$, $E_r$, $S_p$, and $S_r$ are acquired from Eq.~\eqref{eq:E_p}, Eq.~\eqref{eq:E_r}, Eq.~\eqref{eq:S_p} and Eq.~\eqref{eq:S_r}. This integrated loss function adaptively adjusts the learning emphasis based on factor effectiveness: when protective factors show effectiveness ($E_p = 1$), the model prioritizes learning their patterns, similarly for risk factors ($E_r = 1$).

\subsubsection{Suicide Risk Prediction}
To assess the subsequent suicide risk level for each temporal sequence, we process the extracted features and produce the final prediction vector as follows:

\begin{equation}
\hat{y}^{sr} = f_8(ReLU(f_9(\mathbf{u})))
\label{eq:suicide risk}
\end{equation}
where $f_8(.)$ and $f_9(.)$ are fully-connected layers. $\mathbf{u}$ is the temporal attention-aware embedding of post sequence acquired from Eq.~\eqref{eq:attention}.

To evaluate subsequent suicide risk levels, we consider the ordinal relationship between different risk categories. Considering the ordinal relationship is crucial for suicide risk assessment because the suicide risk levels form a natural progression from lower to higher severity, where misclassifying a high-risk case as low-risk carries significantly greater consequences than adjacent-level errors. 
Since ordinal regression loss~\cite{DBLP:conf/cvpr/DiazM19} can capture the natural ordering of risk levels while maintaining different penalty weights for various severity of misclassifications, we implement the ordinal regression loss as an objective function for suicide risk prediction. For each true label $k^a$ in the ordered set $\{0: \text{indicator}, 1: \text{ideation}, 2: \text{behavior}, 3: \text{attempt}\}$, we first calculate the absolute distances between $k^a$ and each possible label value $k^i \in \{0,1,2,3\}$ to form a distance vector $\phi = \alpha|k^a - k^i|$, where $\alpha$ is a penalty parameter for wrong predictions. Then, we obtain the distance-based probability distribution through $\hat{y}^{sr} = \text{softmax}(-\phi)$. Formally, the distance-based probability distribution of user suicide prediction is represented by:

\begin{equation}
\hat{y}_{j}^{sr} = \frac{e^{-\phi(k^i,k^a)}}{\sum_{j=1}^{L} e^{-\phi(k^i,k^a)}}
\label{eq:ordinal suicide risk labels}
\end{equation}

Finally, we use cross-entropy loss for suicide risk prediction: 
\begin{equation}
\mathcal{L}_{sr} = -\sum_{j=1}^{L} y_{j}^{sr} \log(\hat{y}_{j}^{sr}).
\label{eq:loss function of learning suicide risk}
\end{equation}
where $L$ is the number of suicide risk levels. $\hat{y}_{j}^{sr}$ is the predicted logits.

\subsubsection{Joint Learning}
Our approach addresses different tasks at different granularities: suicide factors recognition on post and suicide risk prediction on dynamic post sequence. To effectively balance these tasks, we adopt uncertainty-weighted loss~\cite{DBLP:conf/cvpr/KendallGC18}. It automatically learns optimal task weights by considering each task's inherent uncertainty, leading to our final objective function:

\begin{equation}
\begin{aligned}
\mathcal{L}_{total} = & \frac{1}{2\sigma_1^2}\mathcal{L}_{sr} + \frac{1}{2\sigma_2^2}\mathcal{L}_{pf} + \frac{1}{2\sigma_3^2}\mathcal{L}_{rf} + \\\
& \frac{1}{2\sigma_4^2}\mathcal{L}_{df} + \log(\sigma_1\sigma_2\sigma_3\sigma_4)
\end{aligned}
\end{equation}
where $\sigma_1$,$\sigma_2$,$\sigma_3$,$\sigma_4$ are learnable parameters representing uncertainty weights for each task, and $\log(\sigma_1\sigma_2\sigma_3\sigma_4)$ serves as a normalization term. $\mathcal{L}_{sr}$, $\mathcal{L}_{pf}$, $\mathcal{L}_{rf}$, $\mathcal{L}_{df}$ are learned from Eq.~\eqref{eq:loss function of learning suicide risk}, Eq.~\eqref{eq: loss function for learning protective factors}, Eq.~\eqref{eq: loss function for learning risk factors}, Eq.~\eqref{eq: loss function for dynamic factor integration}. 
This joint learning framework learns both risk and protective factors that influence suicide risk transition and predicts subsequent suicide risk given a sequence of posts from social media users. The unified framework not only provides accurate suicide risk predictions but also offers interpretable insights into which factors most strongly influence each user's risk trajectory, supporting both automated screening and clinical decision-making.

\section{Experiments}

We conduct extensive experiments to evaluate the proposed model to answer the following questions:
\begin{itemize}
    \item \textbf{RQ1.} Whether our proposed model produces accurate suicide risk prediction in terms of subsequent suicide risk levels compared with state-of-the-art approaches?

    \item \textbf{RQ2.} Do different components (i.e., \textit{Risk and Protective Factors Learning}, and \textit{Dynamic Factors Influence Learning}) help our model to achieve better performance?
    
    \item \textbf{RQ3.} How do hyperparameters impact our proposed model?

    \item \textbf{RQ4.} Does our proposed model show a reasonable and human-understandable trajectory to the user's mental health status change in a real-world case study?
\end{itemize}

\subsection{Experimental Setup}
We first detail our experimental setup in terms of datasets, baselines, experimental setting, and evaluation metrics for good reproducibility and show fair comparisons. 

\subsubsection{Datasets}
For comprehensive evaluations, we conduct our experiments on our proposed dataset and two widely used benchmark suicidal datasets.
In particular, our proposed \textbf{Protective Factor-Aware Dataset (PFA)} captures users' mental state fluctuations over time through frequent posting patterns, providing a more comprehensive view of psychological dynamics than existing benchmarks. 
In addition to our PFA, we conduct future experiments on \textbf{CSSRS-Suicide}~\cite{gaur2019knowledge} and \textbf{RSD-15K}. CSSRS-Suicide contains annotated Reddit posts from 500 users spanning 9 mental health-related subreddits. Each user profile includes an average of 31.5 posts, offering longitudinal insights into their mental health trajectory and enabling more sophisticated analysis of suicide risk factors. RSD-15K is one of the largest annotated suicide-related datasets, containing 14,613 posts from 1,265 users collected from ``r/SuicideWatch''. We use RSD-15K to test our model's ability to handle large-scale data.
We extended the original CSSRS-Suicide and RSD-15K datasets by supplementing them with risk and protective factors following the same annotation protocol as in our proposed dataset.

\subsubsection{Baselines}
We compare our proposed model with eight baseline models in terms of two categories: I) State-of-the-art suicide prediction models: specialized architectures designed for suicide risk detection on social media with post-sequence understanding capabilities; II) Large-language model (LLM)-based suicide risk detection models: adopts state-of-the-art open-source and commercial large language models for suicide risk prediction through prompt engineering and supervised fine-tuning. 
In particular, 
\begin{itemize}
\item \textbf{SISMO~\cite{sawhney2021ordinal} (I)}: SISMO leverages Longformer to encode individual posts, followed by a bidirectional LSTM layer and an attention mechanism for sequential modeling on the suicide risk detection task.
\item \textbf{STATENet~\cite{sawhney2020time} (I)}: STATENet employs a dual-branch architecture that jointly learns representations from historical tweets and the target tweet for suicide risk detection on given users' posts. 
\item \textbf{TSAML~\cite{lee2023towards} (I)}: TSAML employs Sentence-BERT to encode individual posts, followed by a bidirectional LSTM and temporal attention mechanism to predict the highest suicide risk level.

\end{itemize}

For category (II), we compare the following large language models (LLMs): \textbf{Llama 3-8B}, \textbf{Gemma 2-9B}, \textbf{GPT-3.5}, \textbf{GPT-4}, and \textbf{Claude-3.5-sonnet}. These LLMs, trained on millions of diverse text samples, are optimized for instruction following and conversational tasks with strong reasoning capabilities. We adapt them for suicide prediction using carefully crafted prompts. For Llama 3-8B and Gemma 2-9B, we use LLama-Factory~\cite{zheng2024llamafactory} specifically for suicide risk prediction. 

\subsubsection{Experimental setting}

All experiments are conducted using stratified 5-fold cross-validation, ensuring that users in the test set are entirely disjoint from those in the training set with no overlap. The maximum number of epochs for all methods is set to 200, with an early stopping strategy implemented to terminate model training when the loss ceases to improve. Hyperparameters for all models are optimized through grid search, including learning rate, dropout rate, embedding size, the parameter $\alpha$ for ordinal regression, and $\tau$ for controlling alignment sensitivity. The same data splitting and early stopping strategies are consistently applied across all baseline methods. Given the long-tail distribution observed in suicide factors, we employ random forest feature selection to identify the 8 most relevant factors for each dataset. For the LLM prompt, we try different prompt content and find the best prompt setting throughout 10 rounds of experiments.


\subsubsection{Evaluation Metrics}


Traditional classification metrics such as precision and recall are insufficient for evaluating ordinal suicide risk levels, as they cannot distinguish between systematic over-prediction and under-prediction tendencies. Therefore, following~\cite{gaur2019knowledge}, we adopt specialized evaluation metrics tailored for ordered categorical outcomes to better evaluate model performance in the context of ordinal suicide risk assessment.
In particular, we define False Negatives ($FN$) as instances where the predicted risk level is lower than the actual risk level, capturing the ratio of under-predictions to total predictions. Similarly, False Positives ($FP$) represent instances where the predicted risk level exceeds the actual risk level, measuring the ratio of over-predictions to total predictions. True Positives ($TP$) are defined as cases where the predicted risk level precisely matches the actual risk level, accurately identifying the correct category.
Subsequently, we compute Graded Precision (GP), Graded Recall (GR), and Graded F-Score (FS) using the modified false negative (FN) and false positive (FP) values, as proposed by~\cite{gaur2019knowledge}.
\begin{equation}
\begin{gathered}
GP = \frac{TP}{TP+FP} \quad GR = \frac{TP}{TP+FN} \\ FS = \frac{2 \times GP \times GR}{GP + GR}
\end{gathered}
\end{equation}

\subsection{Subsequent Suicide Risk Prediction Performance (RQ1)}
Our proposed model demonstrates superior performance across all three datasets, consistently outperforming baseline methods as shown in Table~\ref{tab:model-comparison-all-datasets}. On the CSSRS-Suicide dataset, the model achieves optimal performance (FS=0.8911), showing a 0.66\% improvement over the strongest baseline TSAML. On the complex large-scale RSD-15K dataset, the model reaches FS=0.6861, surpassing TSAML by 1.73\%. Particularly impressive is the performance on the Protective Factor-Aware Dataset (FS=0.6617), substantially outperforming second-best performer by 10.17\%. This superior performance stems from the model's Sequential Post Modeling module and Risk and Protective Factors Influence Learning module, enabling effective capture of the complete dynamics of suicide risk transitions.

Large language models exhibit a pronounced conservative bias in suicide risk prediction, tending to achieve higher GR scores but lower GP scores, essentially misclassifying moderate-risk cases as high-risk. While this cautious approach reduces the likelihood of missing high-risk users, it increases false positives. More critically, the low GP scores indicate LLMs' failure to comprehensively consider both protective and risk factors, leading to resource misallocation and reduced clinical utility in real-world applications.

Furthermore, among the state-of-the-art baselines in category I, while SISMO achieved relatively higher GR scores, its inability to capture temporal patterns in sequential posts limited its overall performance. Although STATENet framework models varying time intervals between tweets, it ignores the impact of factors on suicide risk detection tasks, which limits its performance. For TSAML, its multi-task architecture jointly learns suicide risk levels and risk factors from post sequences. However, it doesn't consider protective factors in their model, which leads to overestimating the suicide risk levels of users. This finding highlights the critical importance of incorporating protective factors in suicide risk prediction models to achieve more accurate and balanced risk assessment.

\begin{table*}[ht]
\caption{Performance comparison of different models on three datasets using graded metrics. GP, GR, and FS denote Graded Precision, Graded Recall, and F-Score respectively.}
\label{tab:model-comparison-all-datasets}
\centering
\begin{tabular}{l|ccc|ccc|ccc}
\hline
\multirow{2}{*}{\textbf{Model}} & \multicolumn{3}{c|}{\textbf{CSSRS-Suicide}} & \multicolumn{3}{c|}{\textbf{RSD-15K}} & \multicolumn{3}{c}{\textbf{Our dataset}} \\
\cline{2-10}
 & \textbf{GP} & \textbf{GR} & \textbf{FS} & \textbf{GP} & \textbf{GR} & \textbf{FS} & \textbf{GP} & \textbf{GR} & \textbf{FS} \\
\hline
SISMO & 0.9875 & 0.7932 & 0.8797 & \textbf{0.7121} & 0.6221 & 0.6336 & 0.6849 & 0.5018 & 0.5416 \\
STATENet & 0.6649 & 0.8468 & 0.7449 & 0.5584 & 0.6154 & 0.5855 & 0.4550 & 0.6224 & 0.5247 \\
TSAML & 0.9923 & 0.7979 & 0.8845 & 0.6843 & 0.6540 & 0.6688 & 0.6364 & 0.5000 & 0.5600 \\
\hline
Finetuned Llama 3-8B & 0.9501 & 0.8078 & 0.8732 & 0.6545 & 0.5526 & 0.5992 & 0.5198 & 0.5756 & 0.5463 \\
Finetuned Gemma 2-9B & 0.9714 & 0.8176 & 0.8879 & 0.6371 & 0.5960 & 0.6159 & 0.4947 & 0.4312 & 0.4608 \\
GPT-3.5 & 0.5146 & 0.8492 & 0.6408 & 0.2260 & 0.7359 & 0.3458 & 0.1603 & 0.8094 & 0.2676 \\
GPT-4 & 0.3935 & 0.8551 & 0.5390 & 0.4552 & \textbf{0.7598} & 0.5693 & 0.2408 & 0.7849 & 0.3686 \\
Claude-3.5-sonnet & 0.6110 & \textbf{0.8645} & 0.7160 & 0.4581 & 0.7218 & 0.5605 & 0.2411 & \textbf{0.8646} & 0.3770 \\
\hline
Our model & \textbf{0.9992} & 0.8041 & \textbf{0.8911} & 0.6618 & 0.7123 & \textbf{0.6861} & \textbf{0.8203} & 0.5545 
& \textbf{0.6617} \\
\hline
\end{tabular}
\vspace{-0.2in}
\end{table*}

\subsection{Ablation Study (RQ2)}
We conducted an ablation study to evaluate the contribution of each component in our model. The results in Figure~\ref{fig-ablation study} demonstrate the effectiveness of each component in our model architecture.



\subsubsection{Impact of Risk and Protective Factors Learning}
Protective factors and risk factors are crucial as they have different influence on suicide risk transitions over time. To study the impact of \textit{Risk and Protective Factors Learning}, we remove Risk and Protective Factors Learning Process $L_{rf}$ and $L_{pf}$ in Eq.~\eqref{eq: loss function for learning risk factors} and Eq.~\eqref{eq: loss function for learning protective factors}, respectively. We denote such two variants as \textit{w/o RF} and \textit{w/o PF} and show the results in Figure~\ref{fig-ablation study}. First, when protective factors learning are eliminated (\textit{w/o PF}), the model's performance decreases slightly. Secondly, removing the risk factors learning (\textit{w/o RF}) leads to a significant drop in performance, resulting in a lower FS of 0.6800. These decreases indicate that it is essential to learn both protective factors and risk factors for accurate suicide risk prediction.

\subsubsection{Impact of Dynamic Factors Influence Learning}
\begin{figure}[ht]
    \centering
    \includegraphics[width = 0.48\textwidth]{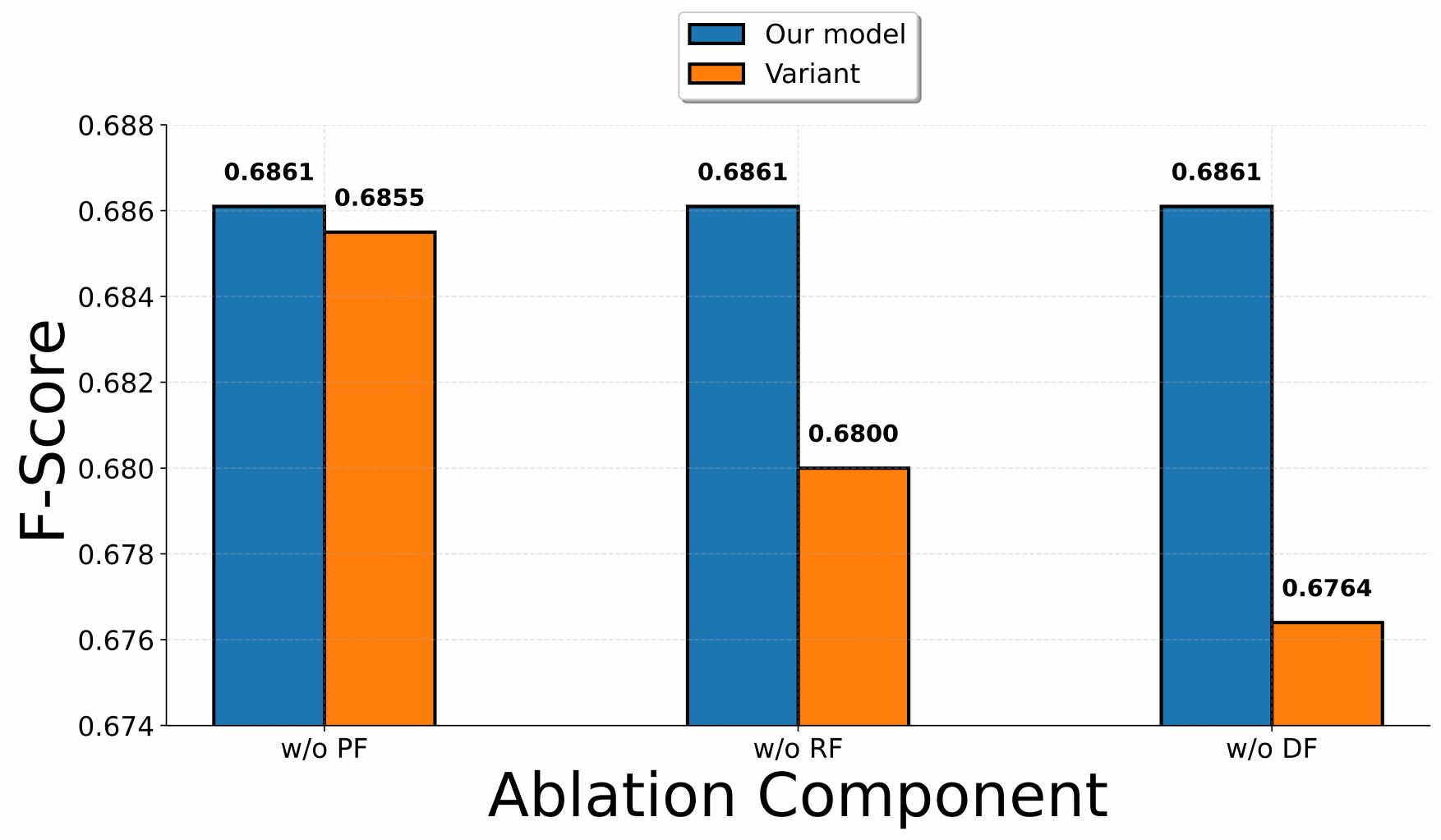}
    \caption{Ablation results over our model components in RSD-15k dataset. ``DF'' denotes dynamic factors influence learning acquired from Eq.~\eqref{eq: loss function for dynamic factor integration}. ``RF'' and ``PF'' indicate risk and protective factors learning respectively, which acquired from Eq.~\eqref{eq: loss function for learning risk factors} and Eq.~\eqref{eq: loss function for learning protective factors}.}
    \label{fig-ablation study}
    \vspace{-0.3in}
\end{figure}

The influence of risk and protective factors demonstrates significant variation across individuals. Our \textit{Dynamic Factors Influence Learning} effectively quantifies the differential impact of these two factor types on future suicide risk transitions, enabling personalized assessment of how these factors contribute to risk escalation or mitigation. To study the impact of this component, we remove the Dynamic Factor Integration $L_{df}$ in Eq.~\eqref{eq: loss function for dynamic factor integration} (\textit{w/o DF}). The result is shown in Figure~\ref{fig-ablation study}. Ablation of the \textit{Dynamic Factors Influence Learning} component results in a substantial performance degradation, with the FS score declining significantly to 0.6764, underscoring the critical contribution of this mechanism to the model's predictive capabilities. These results validate the effectiveness of our model design, where each component contributes meaningfully to the overall performance.

\subsection{Parameter Sensitivity Analysis (RQ3)}\label{sec:section7}
\begin{figure}[!t]
\centering
\subfloat[Varying length of post sequence]{\includegraphics[width=1.6in]{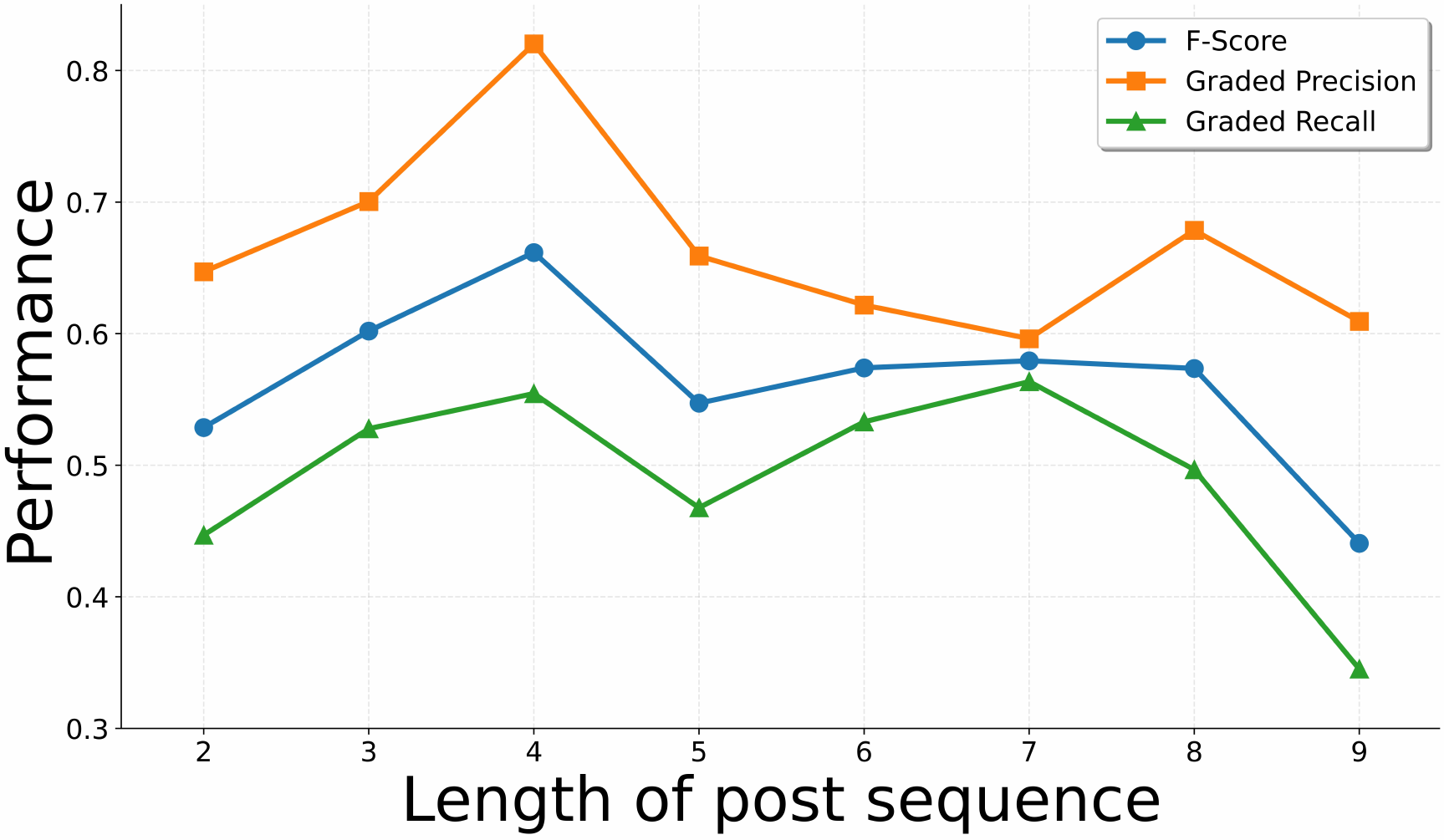}
\label{fig:window_size}}
\hfil
\subfloat[Varying values of $\tau$]{\includegraphics[width=1.6in]{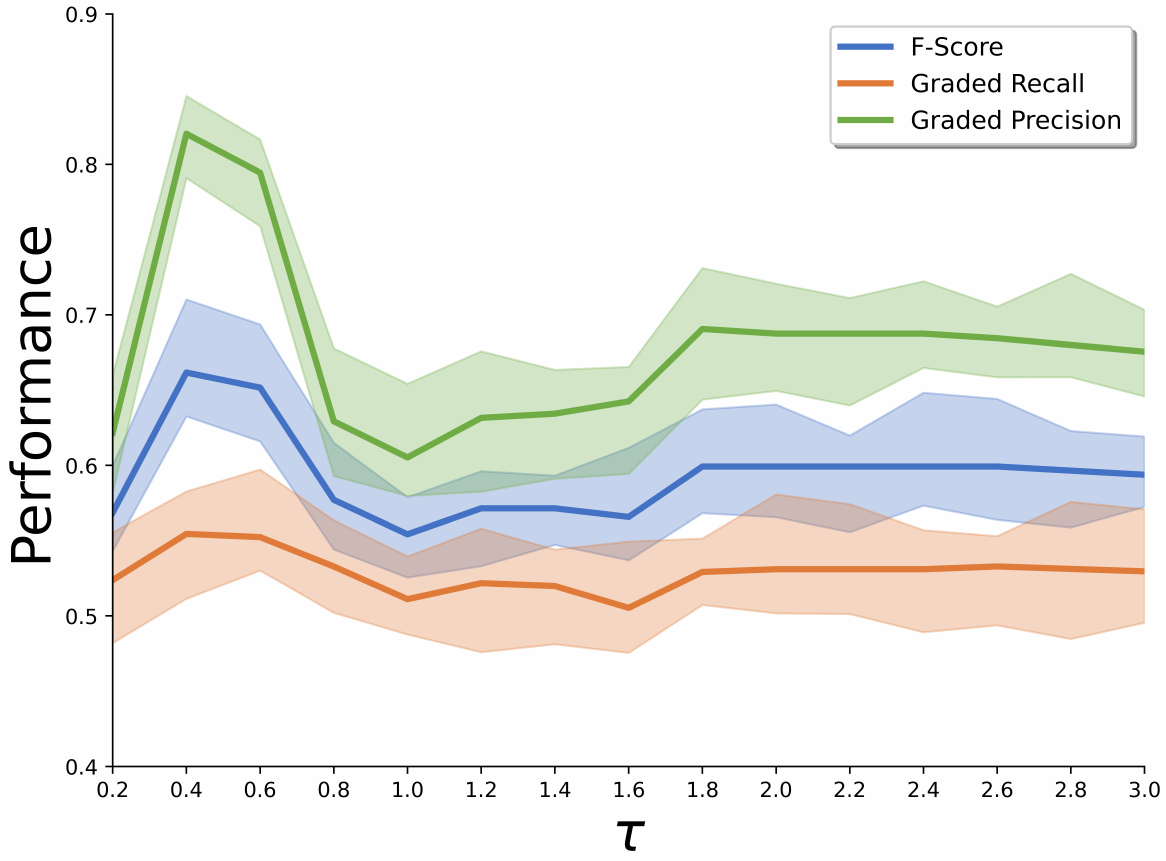}
\label{fig:temperature}}
\caption{F-Score, Graded Precision, Graded Recall of our model against (a) varying the length of post sequence, and (b) varying values of parameter $\tau$ which controls the sensitivity of alignment in Eq.~\eqref{eq:S_p} and Eq.~\eqref{eq:S_r} on the Protective Factor-Aware Dataset (PFA).}
\label{fig_sim}
\vspace{-0.25in}
\end{figure}

We analyze model performance with varying post sequence lengths, starting from 2 posts (the minimum required for our task). In Figure~\ref{fig:window_size}, we can see that F-Score, Graded Precision (GP), Graded Recall (GR) show an increasing trend until reaching length of post sequence 4, where they achieve their peak performance (FS: 0.6617, GP: 0.8203, GR: 0.5545). This suggests that incorporating up to 4 posts in the sequence provides optimal information for the model to make accurate predictions. However, when we further increase the number of posts in user history, both FS and GR metrics begin to decline, indicating that longer sequences may introduce noise or cause the model to lose focus on the most relevant information. This finding indicates that there is an optimal balance between having sufficient context (length of post sequence 4) and avoiding information overload.

Our model has one important parameter $\tau$, which controls the sensitivity of alignment in Eq.~\eqref{eq:S_p} and Eq.~\eqref{eq:S_r}. We conduct a more detailed investigation of this parameter's effect by systematically varying it across the range [0.2, 0.4, ..., 3.0]. In Figure~\ref{fig:temperature}, we observe that when $\tau = 0.2$, our model's performance drastically degrades, which can be attributed to the sharper similarity distribution between factors and user. The model cannot differentiate between correct and wrong predictions as it forcing the model learn like one-hot similarity distribution instead of the soft distribution, resulting in overfitting to a specific factor. When $\tau$ is too large ($\tau \geq 0.8$), the model learns a uniform user-factor similarity distribution, as we heavily degenerates its sensitivity to the critical factors that enable accurate predictions. We observe that $\tau = 0.4$ provides the best model performance by achieving an optimal balance between distribution sharpness and smoothness, allowing the model to maintain sensitivity to critical factors while avoiding overfitting to any single factor pattern.


\subsection{Case Study (RQ4)}
While our framework already incorporates temporal attention mechanisms to capture temporal attention-aware dependencies in user posts, the \textit{Dynamic Factors Influence Learning} serves a fundamentally different and complementary purpose. The temporal attention mechanism primarily focuses on intra-sequence temporal dependencies, learning the relative importance of different posting time steps $T=\{1, ..., t\}$ within a post sequence by addressing \textit{when} certain post content is more critical for understanding mental state evolution. In contrast, the \textit{Dynamic Factors Influence Learning} specifically models cross-modal associations between factors and user's post sequence, addressing \textit{what} extent of both types of factors (i.e., protective factors or risk factors) influence subsequent suicide risk transitions. We illustrate this by a case study as shown in Figure~\ref{case study}:

\begin{figure}[ht]
    \centering
    \includegraphics[width = 0.48\textwidth]{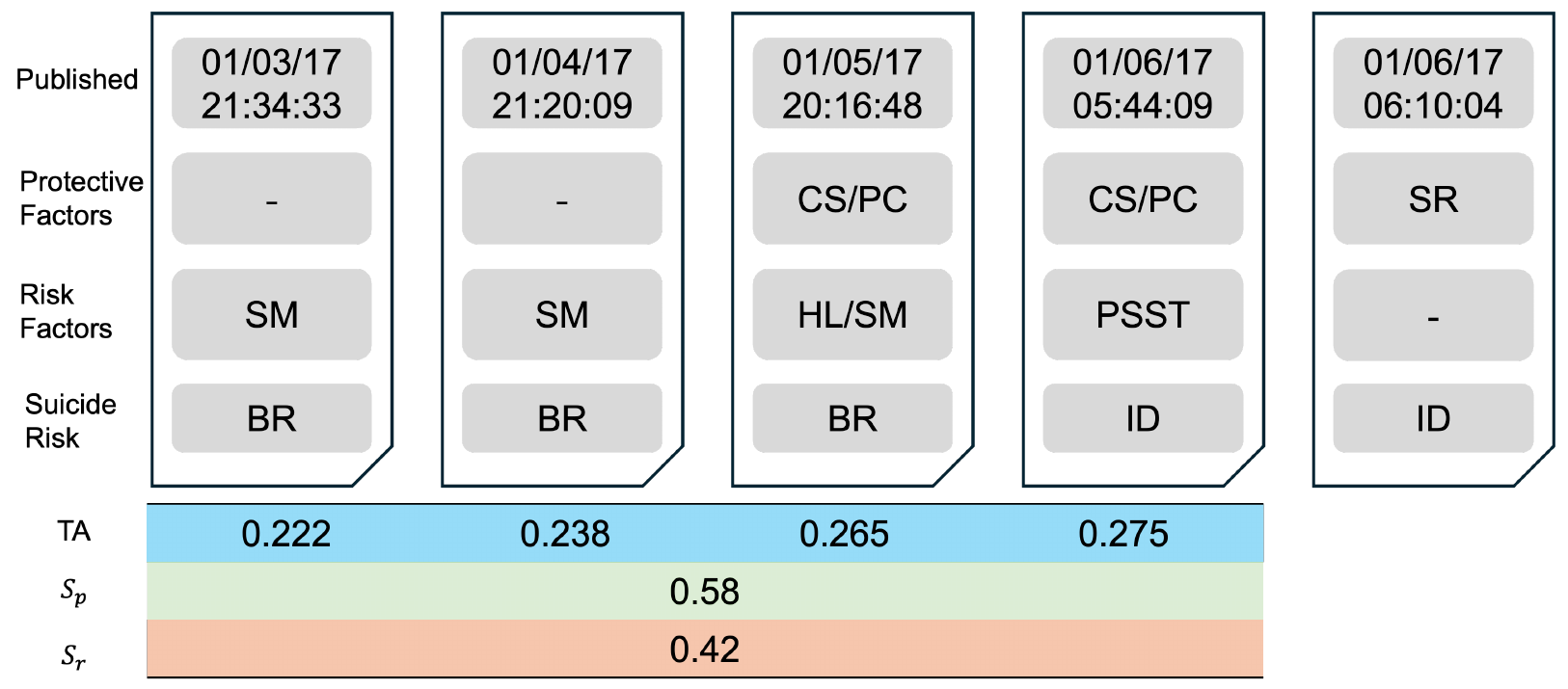}
    \caption{Case study on the influence of factors on suicide risk transition. ``TA'' denotes Temporal attention. $S_p$ and $S_r$ are acquired from Eq.~\eqref{eq:S_p} and Eq.~\eqref{eq:S_r} that offer direct quantification of factors influence strengths for suicide risk transition.}
    \label{case study}
    \vspace{-0.1in}
\end{figure}

Based on the sequence of posts, our model with temporal attention mechanism demonstrates its predictive capability by accurately forecasting the final post's suicide risk level (ID) through the dynamic interplay of protective and risk factors over time. The first two posts (01/03/17-01/04/17) consistently show high suicide risk (BR) with predominant risk factors (SM) and absent protective factors, establishing a baseline of elevated risk. The critical turning point occurs at the third post where protective factors (CS/PC) first emerge. This protective influence becomes increasingly pronounced as evidenced by the fourth post, where the user expressed: ``I would commit suicide before college started, but I decided to give it a chance...'' This statement reflects the emergence of psychological capital and hope, key components of protective factors that facilitate the transition from high-risk (BR) to lower-risk states. The model leverages its temporal attention mechanism to assign greater weight to protective factors in recent posts while simultaneously recognizing that the overall influence from protective factor ($S_p$ = 0.58) substantially exceeds the influence from risk factor ($S_r$ = 0.42), indicating the dominance of protective influences in driving the suicide risk transition. This case study demonstrates the model's ability to track the characteristic rapid suicide risk fluctuations inherent in vulnerable users by learning temporal attention scores and quantifying the influence strength of protective and risk factors that collectively drive potential suicide risk transitions.

\section{Discussion}
\textbf{Ethical Considerations}: The development of suicide risk prediction models presents substantial ethical challenges that demand rigorous attention. Most critically, we emphasize that our model serves as a supplementary tool rather than a substitute for professional clinical evaluation and mental health expertise. Privacy protection forms the cornerstone of our research approach. We established comprehensive de-identification protocols that systematically eliminate all personally identifiable information from collected social media posts. The examples presented in Figure~\ref{fig-example} have been paraphrased to ensure user anonymity. Furthermore, we recognize the potential for unintended applications of our model, highlighting the necessity for responsible deployment within established professional mental health frameworks.

\textbf{Limitation}: Our study faces several important constraints. Although we incorporated mental health professionals from varied backgrounds in both training and annotation phases with cross-validation procedures, suicide risk assessment from social media posts remains fundamentally subjective. Risk indicators can be interpreted differently depending on contextual factors and individual circumstances, introducing inherent variability in judgments.

\section{Conclusion and Future Work}
In this work, we proposed a novel multi-task learning framework that jointly learns (1) risk and protective factors of users on social media over time and (2) subsequent suicide risk. The proposed framework for predicting subsequent suicide risk using dynamic factors influence learning can effectively capture the varying influence of both risk and protective factors on suicide risk transitions. In our experiments, our model outperforms state-of-the-art models in suicide risk detection domain and large language models across three datasets, demonstrating significant improvements in prediction performance. Our model will contribute to advancing suicide prevention research by providing enhanced capabilities for suicide risk assessment, enabling the identification of potentially at-risk individuals before critical transitions occur, ultimately helping to save lives through timely and targeted prevention efforts. Future research will investigate system scalability and practical deployment scenarios.


\newpage

\bibliographystyle{IEEEtran}
\bibliography{references}

\begin{thebibliography}{10}
\providecommand{\url}[1]{#1}
\csname url@samestyle\endcsname
\providecommand{\newblock}{\relax}
\providecommand{\bibinfo}[2]{#2}
\providecommand{\BIBentrySTDinterwordspacing}{\spaceskip=0pt\relax}
\providecommand{\BIBentryALTinterwordstretchfactor}{4}
\providecommand{\BIBentryALTinterwordspacing}{\spaceskip=\fontdimen2\font plus
\BIBentryALTinterwordstretchfactor\fontdimen3\font minus \fontdimen4\font\relax}
\providecommand{\BIBforeignlanguage}[2]{{%
\expandafter\ifx\csname l@#1\endcsname\relax
\typeout{** WARNING: IEEEtran.bst: No hyphenation pattern has been}%
\typeout{** loaded for the language `#1'. Using the pattern for}%
\typeout{** the default language instead.}%
\else
\language=\csname l@#1\endcsname
\fi
#2}}
\providecommand{\BIBdecl}{\relax}
\BIBdecl

\bibitem{fu2007predictive}
K.-w. Fu, K.~Y. Liu, and P.~S. Yip, ``Predictive validity of the chinese version of the adult suicidal ideation questionnaire: psychometric properties and its short version.'' \emph{Psychological Assessment}, vol.~19, no.~4, p. 422, 2007.

\bibitem{scherer2013investigating}
S.~Scherer, J.~Pestian, and L.-P. Morency, ``Investigating the speech characteristics of suicidal adolescents,'' in \emph{2013 IEEE International Conference on Acoustics, Speech and Signal Processing}.\hskip 1em plus 0.5em minus 0.4em\relax IEEE, 2013, pp. 709--713.

\bibitem{karim2020social}
F.~Karim, A.~A. Oyewande, L.~F. Abdalla, R.~C. Ehsanullah, and S.~Khan, ``Social media use and its connection to mental health: a systematic review,'' \emph{Cureus}, vol.~12, no.~6, 2020.

\bibitem{ulvi2022social}
O.~Ulvi, A.~Karamehic-Muratovic, M.~Baghbanzadeh, A.~Bashir, J.~Smith, and U.~Haque, ``Social media use and mental health: a global analysis,'' \emph{Epidemiologia}, vol.~3, no.~1, pp. 11--25, 2022.

\bibitem{pennebaker2001linguistic}
J.~W. Pennebaker, ``Linguistic inquiry and word count: Liwc 2001,'' 2001.

\bibitem{sawhney2018computational}
R.~Sawhney, P.~Manchanda, R.~Singh, and S.~Aggarwal, ``A computational approach to feature extraction for identification of suicidal ideation in tweets,'' in \emph{Proceedings of ACL 2018, Student Research Workshop}, 2018, pp. 91--98.

\bibitem{DBLP:conf/kdd/LeeSJKH23}
D.~Lee, S.~Son, H.~Jeon, S.~Kim, and J.~Han, ``Towards suicide prevention from bipolar disorder with temporal symptom-aware multitask learning,'' in \emph{{KDD}}.\hskip 1em plus 0.5em minus 0.4em\relax {ACM}, 2023, pp. 4357--4369.

\bibitem{johnson2011resilience}
J.~Johnson, A.~M. Wood, P.~Gooding, P.~J. Taylor, and N.~Tarrier, ``Resilience to suicidality: The buffering hypothesis,'' \emph{Clinical psychology review}, vol.~31, no.~4, pp. 563--591, 2011.

\bibitem{rudd2006fluid}
M.~D. Rudd, ``Fluid vulnerability theory: A cognitive approach to understanding the process of acute and chronic suicide risk.'' 2006.

\bibitem{lundman2007psychometric}
B.~Lundman, G.~Strandberg, M.~Eisemann, Y.~Gustafson, and C.~Brulin, ``Psychometric properties of the swedish version of the resilience scale,'' \emph{Scandinavian journal of caring sciences}, vol.~21, no.~2, pp. 229--237, 2007.

\bibitem{du2018extracting}
J.~Du, Y.~Zhang, J.~Luo, Y.~Jia, Q.~Wei, C.~Tao, and H.~Xu, ``Extracting psychiatric stressors for suicide from social media using deep learning,'' \emph{BMC medical informatics and decision making}, vol.~18, pp. 77--87, 2018.

\bibitem{ji2018supervised}
S.~Ji, C.~P. Yu, S.-f. Fung, S.~Pan, and G.~Long, ``Supervised learning for suicidal ideation detection in online user content,'' \emph{Complexity}, vol. 2018, no.~1, p. 6157249, 2018.

\bibitem{venek2017adolescent}
V.~Venek, S.~Scherer, L.-P. Morency, J.~Pestian \emph{et~al.}, ``Adolescent suicidal risk assessment in clinician-patient interaction,'' \emph{IEEE Transactions on Affective Computing}, vol.~8, no.~2, pp. 204--215, 2017.

\bibitem{DBLP:conf/eacl/SawhneyJFS21}
R.~Sawhney, H.~Joshi, L.~Flek, and R.~R. Shah, ``{PHASE:} learning emotional phase-aware representations for suicide ideation detection on social media,'' in \emph{{EACL}}.\hskip 1em plus 0.5em minus 0.4em\relax Association for Computational Linguistics, 2021, pp. 2415--2428.

\bibitem{lee2023towards}
D.~Lee, S.~Son, H.~Jeon, S.~Kim, and J.~Han, ``Towards suicide prevention from bipolar disorder with temporal symptom-aware multitask learning,'' in \emph{Proceedings of the 29th ACM SIGKDD conference on knowledge discovery and data mining}, 2023, pp. 4357--4369.

\bibitem{cao2021learning}
L.~Cao, H.~Zhang, X.~Wang, and L.~Feng, ``Learning users inner thoughts and emotion changes for social media based suicide risk detection,'' \emph{IEEE Transactions on Affective Computing}, 2021.

\bibitem{li2022suicide}
J.~Li, X.~Chen, Z.~Lin, K.~Yang, H.~V. Leong, N.~X. Yu, and Q.~Li, ``Suicide risk level prediction and suicide trigger detection: A benchmark dataset,'' \emph{HKIE Transactions Hong Kong Institution of Engineers}, vol.~29, no.~4, pp. 268--282, 2022.

\bibitem{hawton2007restricting}
K.~Hawton, ``Restricting access to methods of suicide: Rationale and evaluation of this approach to suicide prevention,'' \emph{Crisis}, vol.~28, no.~S1, pp. 4--9, 2007.

\bibitem{rudd2006warning}
M.~D. Rudd, A.~L. Berman, T.~E. Joiner~Jr, M.~K. Nock, M.~M. Silverman, M.~Mandrusiak, K.~Van~Orden, and T.~Witte, ``Warning signs for suicide: Theory, research, and clinical applications,'' \emph{Suicide and Life-Threatening Behavior}, vol.~36, no.~3, pp. 255--262, 2006.

\bibitem{wang2022suicide}
X.~Wang, Z.~Lu, and C.~Dong, ``Suicide resilience: A concept analysis,'' \emph{Frontiers in psychiatry}, vol.~13, p. 984922, 2022.

\bibitem{posner2011columbia}
K.~Posner, G.~K. Brown, B.~Stanley, D.~A. Brent, K.~V. Yershova, M.~A. Oquendo, G.~W. Currier, G.~A. Melvin, L.~Greenhill, S.~Shen \emph{et~al.}, ``The columbia--suicide severity rating scale: initial validity and internal consistency findings from three multisite studies with adolescents and adults,'' \emph{American journal of psychiatry}, vol. 168, no.~12, pp. 1266--1277, 2011.

\bibitem{fleiss1971measuring}
J.~L. Fleiss, ``Measuring nominal scale agreement among many raters.'' \emph{Psychological bulletin}, vol.~76, no.~5, p. 378, 1971.

\bibitem{gaur2019knowledge}
M.~Gaur, A.~Alambo, J.~P. Sain, U.~Kursuncu, K.~Thirunarayan, R.~Kavuluru, A.~Sheth, R.~Welton, and J.~Pathak, ``Knowledge-aware assessment of severity of suicide risk for early intervention,'' in \emph{The world wide web conference}, 2019, pp. 514--525.

\bibitem{zirikly2019clpsych}
A.~Zirikly, P.~Resnik, O.~Uzuner, and K.~Hollingshead, ``Clpsych 2019 shared task: Predicting the degree of suicide risk in reddit posts,'' in \emph{Proceedings of the sixth workshop on computational linguistics and clinical psychology}, 2019, pp. 24--33.

\bibitem{amare2018prevalence}
T.~Amare, S.~Meseret~Woldeyhannes, K.~Haile, and T.~Yeneabat, ``Prevalence and associated factors of suicide ideation and attempt among adolescent high school students in dangila town, northwest ethiopia,'' \emph{Psychiatry journal}, vol. 2018, no.~1, p. 7631453, 2018.

\bibitem{amiri2022prevalence}
S.~Amiri, ``Prevalence of suicide in immigrants/refugees: a systematic review and meta-analysis,'' \emph{Archives of suicide research}, vol.~26, no.~2, pp. 370--405, 2022.

\bibitem{harris2019factors}
K.~Harris, P.~Gooding, G.~Haddock, and S.~Peters, ``Factors that contribute to psychological resilience to suicidal thoughts and behaviours in people with schizophrenia diagnoses: qualitative study,'' \emph{BJPsych open}, vol.~5, no.~5, p. e79, 2019.

\bibitem{henderson2015responses}
A.~R. Henderson and A.~Cock, ``The responses of young people to their experiences of first-episode psychosis: harnessing resilience,'' \emph{Community mental health journal}, vol.~51, pp. 322--328, 2015.

\bibitem{o2014psychology}
R.~C. O'Connor and M.~K. Nock, ``The psychology of suicidal behaviour,'' \emph{The Lancet Psychiatry}, vol.~1, no.~1, pp. 73--85, 2014.

\bibitem{bryan2020nonlinear}
C.~J. Bryan, J.~E. Butner, A.~M. May, K.~F. Rugo, J.~A. Harris, D.~N. Oakey, D.~C. Rozek, and A.~O. Bryan, ``Nonlinear change processes and the emergence of suicidal behavior: A conceptual model based on the fluid vulnerability theory of suicide,'' \emph{New ideas in psychology}, vol.~57, p. 100758, 2020.

\bibitem{de2013predicting}
M.~De~Choudhury, M.~Gamon, S.~Counts, and E.~Horvitz, ``Predicting depression via social media,'' in \emph{Proceedings of the international AAAI conference on web and social media}, vol.~7, no.~1, 2013, pp. 128--137.

\bibitem{reimers2019sentence}
N.~Reimers, ``Sentence-bert: Sentence embeddings using siamese bert-networks,'' \emph{arXiv preprint arXiv:1908.10084}, 2019.

\bibitem{azim2022detecting}
T.~Azim, L.~G. Singh, and S.~E. Middleton, ``Detecting moments of change and suicidal risks in longitudinal user texts using multi-task learning,'' in \emph{Proceedings of the Eighth Workshop on Computational Linguistics and Clinical Psychology}, 2022, pp. 213--218.

\bibitem{oliffe2012you}
J.~L. Oliffe, J.~S. Ogrodniczuk, J.~L. Bottorff, J.~L. Johnson, and K.~Hoyak, ``“you feel like you can’t live anymore”: Suicide from the perspectives of canadian men who experience depression,'' \emph{Social science \& medicine}, vol.~74, no.~4, pp. 506--514, 2012.

\bibitem{hochreiter1997long}
S.~Hochreiter and J.~Schmidhuber, ``Long short-term memory,'' \emph{Neural computation}, vol.~9, no.~8, pp. 1735--1780, 1997.

\bibitem{DBLP:conf/cvpr/DiazM19}
\BIBentryALTinterwordspacing
R.~Diaz and A.~Marathe, ``Soft labels for ordinal regression,'' in \emph{{IEEE} Conference on Computer Vision and Pattern Recognition, {CVPR} 2019, Long Beach, CA, USA, June 16-20, 2019}.\hskip 1em plus 0.5em minus 0.4em\relax Computer Vision Foundation / {IEEE}, 2019, pp. 4738--4747. [Online]. Available: \url{http://openaccess.thecvf.com/content\_CVPR\_2019/html/Diaz\_Soft\_Labels\_for\_Ordinal\_Regression\_CVPR\_2019\_paper.html}
\BIBentrySTDinterwordspacing

\bibitem{DBLP:conf/cvpr/KendallGC18}
\BIBentryALTinterwordspacing
A.~Kendall, Y.~Gal, and R.~Cipolla, ``Multi-task learning using uncertainty to weigh losses for scene geometry and semantics,'' in \emph{2018 {IEEE} Conference on Computer Vision and Pattern Recognition, {CVPR} 2018, Salt Lake City, UT, USA, June 18-22, 2018}.\hskip 1em plus 0.5em minus 0.4em\relax Computer Vision Foundation / {IEEE} Computer Society, 2018, pp. 7482--7491. [Online]. Available: \url{http://openaccess.thecvf.com/content\_cvpr\_2018/html/Kendall\_Multi-Task\_Learning\_Using\_CVPR\_2018\_paper.html}
\BIBentrySTDinterwordspacing

\bibitem{sawhney2021ordinal}
R.~Sawhney, H.~Joshi, S.~Gandhi, and R.~R. Shah, ``Towards ordinal suicide ideation detectionon social media,'' in \emph{Proceedings of 14th ACM International Conference On Web Search And Data Mining}, ser. WSDM '21.\hskip 1em plus 0.5em minus 0.4em\relax New York, NY, USA: Association for Computing Machinery, Mar. 2021.

\bibitem{sawhney2020time}
R.~Sawhney, H.~Joshi, S.~Gandhi, and R.~Shah, ``A time-aware transformer based model for suicide ideation detection on social media,'' in \emph{Proceedings of the 2020 conference on empirical methods in natural language processing (EMNLP)}, 2020, pp. 7685--7697.

\bibitem{zheng2024llamafactory}
\BIBentryALTinterwordspacing
Y.~Zheng, R.~Zhang, J.~Zhang, Y.~Ye, Z.~Luo, Z.~Feng, and Y.~Ma, ``Llamafactory: Unified efficient fine-tuning of 100+ language models,'' in \emph{Proceedings of the 62nd Annual Meeting of the Association for Computational Linguistics (Volume 3: System Demonstrations)}.\hskip 1em plus 0.5em minus 0.4em\relax Bangkok, Thailand: Association for Computational Linguistics, 2024. [Online]. Available: \url{http://arxiv.org/abs/2403.13372}
\BIBentrySTDinterwordspacing

\end{thebibliography}





\newpage

\end{document}